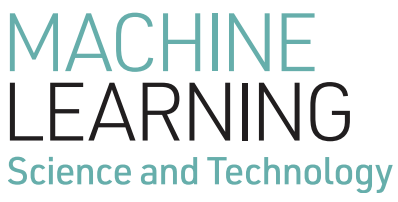

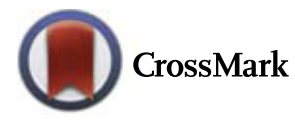

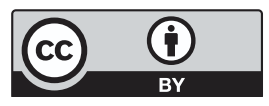

PAPER

# Training multi-layer binary neural networks with random local binary error signals

**Luca Colombo***, **Fabrizio Pittorino** and **Manuel Roveri**

Department of Electronics, Information and Bioengineering, Politecnico di Milano, Via Ponzio 34/5, 20133 Milano, Italy

* Author to whom any correspondence should be addressed.

**E-mail:** luca2.colombo@polimi.it, fabrizio.pittorino@polimi.it and manuel.roveri@polimi.it

## Abstract

Binary neural networks (BNNs) significantly reduce computational complexity and memory usage in machine and deep learning by representing weights and activations with just one bit. However, most existing training algorithms for BNNs rely on quantization-aware floating-point stochastic gradient descent (SGD), limiting the full exploitation of binary operations to the inference phase only. In this work, we propose, for the first time, a fully binary and gradient-free training algorithm for multi-layer BNNs, eliminating the need for back-propagated floating-point gradients. Specifically, the proposed algorithm relies on local binary error signals and binary weight updates, employing integer-valued hidden weights that serve as a synaptic metaplasticity mechanism, thereby enhancing its neurobiological plausibility. Our proposed solution enables the training of binary multi-layer perceptrons by using exclusively XNOR, Popcount, and increment/decrement operations. Experimental results on multi-class classification benchmarks show test accuracy improvements of up to +35.47% over the only existing fully binary single-layer state-of-the-art solution. Compared to full-precision SGD, our solution improves test accuracy by up to +35.30% under the same total memory demand, while also reducing computational cost by two to three orders of magnitude in terms of the total number of Boolean gates. The proposed algorithm is made available to the scientific community as a public repository.

## 1. Introduction

In recent years, machine learning (ML) and deep learning (DL) models have become more and more accurate in solving increasingly challenging tasks. However, this improvement has come at the cost of significantly larger model sizes. The growing number of parameters has led to greater computational demands, increased memory usage, and higher energy consumption [1, 2]. A promising approach that tries to overcome this drawback is represented by quantization. This technique aims at reducing the precision of model weights and/or activations from floating-point values to low-bit integers, resulting in model compression and faster execution [3, 4].

Binary neural networks (BNNs) represent the simplest and most extreme form of low-bit quantized models by constraining weights and activations to 1-bit values (typically $\pm 1$). This design drastically reduces memory footprint and enables highly efficient bitwise operations, such as XNOR and Popcount, to replace multiplications and additions. As a result, BNNs significantly lower computational complexity and energy consumption compared to their full-precision counterparts [5, 6]. The literature in this field, however, mainly introduces solutions that rely on floating-point quantization-aware training (QAT). In QAT, weights are binarized only in the forward pass, while floating-point gradients computed using full-precision stochastic gradient descent (SGD) optimize the model parameters during the backward pass. This prevents the exploitation of 1-bit operations during training [7–9]. Therefore, while a vast body of literature exists on BNNs, a critical limitation persists: nearly all methods rely on full-precision backpropagation or maintain full-precision hidden variables. To the best of our knowledge, only the work by Baldassi *et al* [10] presents an alternative fully binary learning algorithm. However, it is limited to single hidden layer BNNs and cannot be used to train deeper architectures.

In this perspective, the paper aims to address the following research question: *Is it possible to train multi-layer BNNs relying only on binary error signals and binary updates?* To the best of our knowledge, we propose, for the first time in the literature, a novel fully binary and gradient-free learning algorithm capable of effectively and efficiently training binary multi-layer perceptrons (BMLPs) without relying on full-precision backpropagation. Specifically, the proposed algorithm directly optimizes the non-differentiable 01-Loss function, thereby eliminating the need for differentiable surrogate losses (e.g. cross-entropy) and bypassing the straight-through estimator [7]. As a result, binary weights can be optimized directly within their natural discrete domain, the hypercube $\{\pm 1\}^N$ [6]. Our algorithm leverages random local binary error signals, generated by fixed random classifiers at each layer. These signals drive updates to integer-valued hidden metaplastic weights using only XNOR, Popcount, and increment/decrement operations. Crucially, *all* values—inputs, activations, visible weights, and weight updates—are constrained to 1-bit representations. The integer-valued hidden weights act as a synaptic metaplasticity mechanism, encoding the confidence associated with each visible binary weight. This mechanism helps mitigate catastrophic forgetting [11]. In doing so, our algorithm aligns with the broader class of neurobiologically plausible local-learning methods [12–15], while pushing it a step forward by enforcing *fully binary forward and backward* passes.

In summary, our contributions are:

1. A BMLP architecture with fixed and random binary local classifiers. This ensures *binary*, *local*, and *randomized* error signals while enabling independent layer-wise training.
2. A multi-layer learning algorithm specifically designed to train the proposed BMLP, relying exclusively on *binary error signals*.
3. Binary error signals, in turn, enable the design of a *gradient-free* training algorithm that relies exclusively on *binary updates* to integer-valued hidden metaplastic weights.

Consequently, the proposed solution: *(i)* eliminates the traditional full-precision SGD algorithm for computing floating-point gradients and weight updates, reducing memory footprint; *(ii)* enables deep BMLP learning by training multiple layers independently, paving the way for binary training of state-of-the-art (SotA) models; and *(iii)* allows for the exclusive use of efficient bitwise operations even during the *learning phase*, drastically reducing both computational complexity and execution time.

By establishing a feasible method for fully binary multi-layer training, this work lays the foundation for a new class of highly optimized learning algorithms. The proposed training method opens new possibilities for training BNNs in extremely constrained environments, where conventional training algorithms that rely on back-propagated gradients and floating-point arithmetic may be infeasible. These scenarios include Tiny ML (TinyML), privacy-preserving ML (PPML), and neuromorphic or spiking systems, as discussed in detail in section 6.

Experimental evaluations on multi-class classification benchmarks show test accuracy improvements of up to +35.47% over the fully binary single-layer SotA algorithm [10]. Compared to full-precision SGD, our solution improves test accuracy by up to +35.30% under the same total memory demand—including the model, activations, and input dataset. Additionally, it reduces computational cost by two to three orders of magnitude in terms of the total number of Boolean gates required for model training.

The paper is organized as follows. Section 2 surveys related research in BNNs and neurobiologically inspired local learning rules. Section 3 details our fully binary and gradient-free training algorithm. Experimental results are presented in section 4. Section 5 discusses the scope and current limitations of the work, while section 6 concludes the paper and outlines directions for future research.

## 2. Related literature

This section reviews the related literature. Specifically, section 2.1 discusses BNN solutions that use binary forward passes but rely on full-precision SGD in the backward pass. In section 2.2, BNN approaches that enable both binary forward and binary backward passes are presented. Section 2.3 examines a stream of literature on neurobiologically plausible algorithms, positioning our algorithm within this context. Lastly, section 2.4 summarizes our contributions relative to the presented literature.

### 2.1. Binary forward, floating-point backward

Most of the literature falls into the first category, comprising QAT-based approaches. In QAT, BNNs employ 1-bit weights and activations only during the forward pass, but rely on full-precision SGD for computing floating-point gradients and weight updates during the backward pass. The seminal work by Courbariaux

**Table 1.** Comparison with existing solutions.

| Algorithm | Binary forward | Binary backward | Multi-layer architectures |
|---|---|---|---|
| QAT [7, 8, 16–22, 24–27] | Yes[a] | No | Yes |
| Baldassi *et al* [10] | Yes | Yes | No |
| Our proposed solution | **Yes** | **Yes** | **Yes** |

[a] exploit full-precision input and/or output layers or scaling factors.

*et al* [7] introduced weights and activations binarization using the *sign* function, replacing most arithmetic operations in deep neural networks with bitwise operations. Building on this, XNOR-Net [8] incorporated per-channel floating-point scaling factors to reduce binarization error. Further refinements introduced more sophisticated techniques. ABC-Net [16] approximated full-precision weights using a linear combination of multiple binary weight bases and employed multiple binary activations to mitigate information loss. Bi-Real Net [17] introduced shortcuts in deep binary convolutional neural networks (CNNs) to narrow the accuracy gap between 1-bit and floating-point models. Lastly, AdaBin [18] and Schiavone *et al* [19] replaced the fixed set $\{\pm 1\}$ with optimizable binary sets $\{\pm\alpha\}$, where $\alpha \in \mathbb{R}$, allowing each layer to learn suitable weight and activation representations.

Recent advancements in BNN training have also focused on both optimization theory and practical deployment on resource-constrained hardware. On the theoretical side, works such as [20] questioned the necessity of latent real-valued weights in BNN optimization, proposing alternative training paradigms. Nevertheless, these approaches still rely on full-precision gradient computation. Concurrently, a growing body of research addresses the feasibility of training BNNs directly on edge devices. For instance, [21] and [22] propose memory-efficient algorithms and system-level frameworks to mitigate the memory limitations inherent in on-device learning. These BNN-specific methods are part of a broader movement toward low-memory DL [23]. Overall, this literature underscores a dual objective: enhancing both the training efficiency and deployability of BNNs in real-world, resource-constrained settings.

Despite these advances, all the aforementioned methods, along with others [24–27], rely on full-precision SGD to compute gradient updates used for optimizing floating-point model parameters, as summarized in table 1. Consequently, the memory and computational advantages of BNNs are lost during training. Moreover, most modern approaches retain floating-point parameters in the first or last layer, or incorporate learned floating-point scaling factors to mitigate accuracy degradation [16, 17]. In contrast, our proposed algorithm eliminates all full-precision components. Specifically, we introduce a gradient-free algorithm that enables fully binarized forward and backward passes for training BMLPs.

### 2.2. Binary forward, binary backward

The second category includes works that enable both binary forward and binary backward passes. To the best of our knowledge, the only existing work in this category is by Baldassi *et al* [10], who proposed a single-layer training algorithm based on the Clipped Perceptron with Reinforcement (CP+R) rule. Their approach employs a custom and fixed NN architecture consisting of three components: *(i)* the fully-connected layer being trained, *(ii)* a custom sparse grouping layer that clusters perceptrons from the previous layer, and *(iii)* a final fully-connected random classifier. The training algorithm is specifically tailored to this specific architecture. It leverages integer-valued hidden variables for gradient-free updates, aligning with research on neurobiologically plausible single-neuron learning algorithms [28–33]. While Baldassi *et al* [10] present an interesting approach, its solution cannot be directly extended to multi-layer BNNs due to the rigidity of both the NN architecture and the training algorithm.

Conversely, our proposed solution overcomes these limitations by relaxing NN architectural constraints and enabling the training of *multi-layer* BNNs with an arbitrary number of hidden layers. As highlighted in table 1, ours is the first solution that provides a generalized fully binary learning algorithm able to train multi-layer BNNs without relying on floating-point gradient computations.

### 2.3. Neurobiological plausibility and local error signals

Beyond research on BNNs, our proposed solution aligns with a broader literature advocating for *local* or *randomized* learning signals as more biologically plausible alternatives to backpropagation. We emphasize that, unlike our approach, these solutions rely on floating-point forward and floating-point backward passes. Feedback alignment (FA) [12] and direct FA [13] address the weight transport problem by using fixed random matrices. This strategy allows error signals to be back-propagated without requiring exact weight symmetry, achieving near-SGD accuracy on image classification benchmarks. Local error methods [34–37] introduce auxiliary classifiers and local loss functions at each layer, enabling independent layer-wise training

**Table 2.** Memory footprint and computational cost comparison between our proposed solution and existing literature. We consider a layer of size $M \times M$. Computational cost is expressed in terms of number of Boolean gate operations, considering a single input sample.

(a) Memory footprint.

| Algorithm | Weights | Activations | Weight updates |
|---|---|---|---|
| QAT (with SGD) | $32M^2$ | $M^{a}$ | $32M^2$ |
| Baldassi *et al* [10] | $8M^2$ | $M$ | $M^2$ |
| Our proposed solution | $8M^2$ | $M$ | $M^2$ |

(b) Computational cost.

| Algorithm | Forward | Backward | Weight update |
|---|---|---|---|
| QAT (with SGD) | $M^2 + 10M^{2\dagger}$ | $10^4M^2 + 10^4(M-1)^{\dagger}$ | $10^4M^2 + 10^4M^{2\dagger}$ |
| Baldassi *et al* [10] | $M^2 + 10M^2$ | — | $M^2 + 10M^2 + 10M^2$ |
| Our proposed solution | $M^2 + 10M^2$ | — | $M^2 + 10M^2 + 10M^2$ |

[a] most QAT approaches employ full-precision scaling factors, which are omitted for simplicity.

without a global backward pass. Similarly, direct random target projection [14] treats one-hot labels as direct proxies for error signals, eliminating explicit backpropagation altogether.

These solutions align with neurobiological constraints by avoiding exact weight symmetry and global error transport. They also reduce memory overhead by eliminating the need to store intermediate activations for end-to-end backpropagation. However, they still operate entirely in the full-precision domain, using floating-point weights and activations, computing floating-point local errors, and back-propagating floating-point gradients through trainable auxiliary classifiers. In contrast, our proposed method extends local learning and randomized error signal paradigms by employing random binary classifiers as local loss evaluators at each layer. This design allows the training algorithm to rely solely on random binary local error signals, which in turn drive binary increment/decrement operations on integer-valued metaplastic weights [30, 31]. As a result, our method introduces both binary local error signals and binary weight updates into BNN training, further bridging the gap between resource-efficient learning algorithms and biologically inspired credit assignment rules.

### 2.4. Contributions of the proposed solution

In summary, most BNN research still relies on QAT, which uses floating-point backward passes. The only fully binary training method [10], on the other hand, is limited to single-layer BNNs and cannot scale to deeper architectures. Our proposed algorithm addresses this gap by generalizing fully binary training to deep and arbitrarily large BNNs. Specifically, it trains BMLPs using random local binary error signals, which in turn drive purely binary increment/decrement updates on integer-valued weights. As a result, the algorithm completely eliminates the need for floating-point gradient computation and backpropagation. In tables 2(a) and (b), we compare memory footprint and computational cost of our proposed solution with those of existing methods. Compared to QAT-based methods, our approach achieves a $4\times$ reduction in memory usage for hidden weights and a $32\times$ reduction for weight updates. In terms of computation, we observe a significant $10^3\times$ reduction in the number of Boolean gate operations required for weight updates. A more detailed analysis of memory and computational requirements is provided in sections 3.5 and 4.2.

Hence, our work contributes to two key areas. First, it advances the BNN field by providing the first multi-layer training algorithm with fully binarized forward and backward passes. Second, it contributes to the bio-inspired learning literature by introducing a learning rule based local and random binary error signals.

## 3. Proposed solution

This section presents our proposed fully binary and gradient-free algorithm for training BMLPs. Specifically, the proposed algorithm, as illustrated in figure 1(b), operates at three different levels:

1. The *network* level, where the forward pass takes place and the local loss functions are computed to identify layers committing an error.
2. The *layer* level, operating independently for each layer identified at the previous step, where the perceptrons to be updated are selected.
3. The *perceptron* level, operating independently for each layer and for each previously identified perceptron, where the weights are updated.

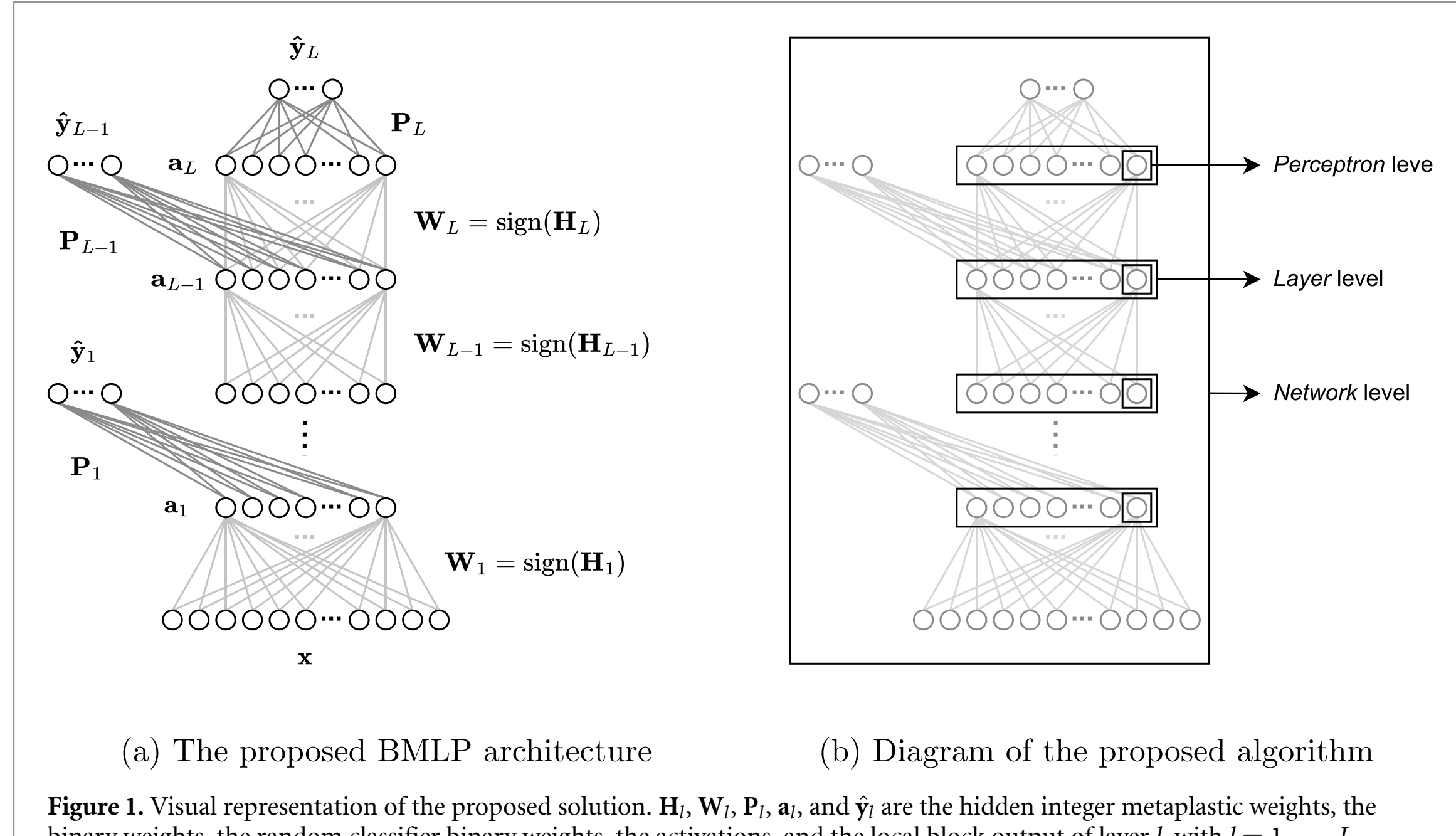

(a) The proposed BMLP architecture (b) Diagram of the proposed algorithm

**Figure 1.** Visual representation of the proposed solution. $\mathbf{H}_l$, $\mathbf{W}_l$, $\mathbf{P}_l$, $\mathbf{a}_l$, and $\hat{\mathbf{y}}_l$ are the hidden integer metaplastic weights, the binary weights, the random classifier binary weights, the activations, and the local block output of layer $l$, with $l = 1, \ldots, L$, respectively. $\mathbf{x}$ is the binary input.

Section 3.1 details the architecture of the BMLP under consideration. Sections 3.2–3.4 explain the proposed algorithm at the network, layer, and perceptron level, respectively. Lastly, section 3.5 provides an overview of the computational costs and memory footprint of the proposed solution.

### 3.1. BNN architecture

The proposed fully binary and gradient-free learning algorithm is designed for the training of BMLPs on multi-class classification problems defined on a dataset $\mathbf{X} = \{\mathbf{x}^\mu, y^\mu\}_{\mu=1}^{N}$, where $\mathbf{x}^\mu \in \mathbb{R}^{K_0}$ are the input patterns of dimension $K_0$, $y^\mu \in \{1, \ldots, c\}$ their corresponding labels (where $c$ is the number of classes), and $N$ is the size of the training set.

The proposed BMLP architecture is shown in figure 1(a). Let $L$ be the total number of fully-connected layers within the considered BMLP, and let $K_l$ be the dimension of layer $l \in \{1, \ldots, L\}$. It should be noted that the input dimension of the BMLP is $K_0$, while the output dimension is $c$. Inspired by [31], each layer $l$ is represented by two matrices: the hidden metaplastic integer weights $\mathbf{H}_l$ of size $K_{l-1} \times K_l$, which are updated during the backward pass, and the binary weights $\mathbf{W}_l$, where $\mathbf{W}_l = \mathit{sign}(\mathbf{H}_l)$, that are used in the forward pass. The hidden metaplastic variables $h \in \mathbf{H}_l$ can be interpreted as the *confidence* of each binary weight in assuming its current value, as the larger the absolute value of $h$, the more difficult for the variable $w$ to change its sign.

Each fully-connected layer $l$ is also associated with a fixed random classifier $\mathbf{P}_l$ of size $K_l \times c$, which is a fully-connected layer with weights $\rho$ uniformly drawn from $\{\pm 1\}$. The objective of these random classifiers is to reduce the dimension of the activations $\mathbf{a}_l$ of the $l$th layer to match the output dimension $c$ and produce the local output $\hat{\mathbf{y}}_l$, which is used during the backward pass by the local loss function $\mathcal{L}$, along with the true label $y$, to compute the local error $\ell_l$. This local error is then used by the training algorithm to compute the weight updates of layer $l$. In the last layer (i.e. $l = L$), the random classifier $\mathbf{P}_L$ serves as the final output layer, where the network output $\hat{\mathbf{y}}_L$ is produced and the final accuracy is evaluated. Once the training phase is completed, the intermediate random classifiers $\mathbf{P}_l$, with $l \in \{1, \ldots, L-1\}$, can either be discarded or employed for early exit strategies [38–40]. We emphasize that, if all layers $L$ share the same dimension $K_l$, a unique random classifier $\mathbf{P}$ can be considered for the training phase.

It is worth noting that the proposed architecture enables the independent training of each layer and allows the parallelization of the training procedure. Moreover, it allows the binary training of arbitrarily deep BMLPs, as explained in the following sections.

### 3.2. Network-level: forward pass, robustness and local loss

The first step of the proposed fully binary and gradient-free learning algorithm operates at the network level. Specifically, as detailed in algorithm 1, it works as follows. First, for each layer $l$, the hidden metaplastic weights $\mathbf{H}_l$ and the random classifier weights $\mathbf{P}_l$ are initialized with random weights uniformly sampled from

**Algorithm 1.** Proposed network-level BMLP training algorithm.

**Data: X**: training data
**Variables:** $L$: number of layers, $K_l$: number of perceptrons in layer $l$, $\mathbf{H}_l$: hidden metaplastic weights in layer $l$, $\mathbf{P}_l$: random classifier weights associated to layer $l$, $E^e$: fraction of training errors at epoch $e$
**Hyperparameters:** $e$: number of epochs, $bs$: mini-batch size, $r$: robustness, $\gamma$: group size, $p_r$: reinforcement probability

1 **def** *NetworkUpdate*$(\mathbf{X}, L, e, bs, r, \gamma, p_r)$**:**
2 | Initialize $\{\mathbf{H}_l, \mathbf{P}_l\}$ **foreach** *layer* $l = 1, \ldots, L$
3 | **foreach** *epoch e* **do**
4 | | **foreach** $(\mathbf{x}, \mathbf{y}) \subseteq \mathbf{X}$, $|\mathbf{x}| = |\mathbf{y}| = bs$ **do**
5 | | | Binarize input data $\mathbf{a}_0 = MedianBinarization(\mathbf{x})$
6 | | | **foreach** $l = 1, \ldots, L$ **do**
7 | | | | Initialize set of pattern indexes to update $\mathcal{M}_l \leftarrow \emptyset$
8 | | | | Compute pre-activations $\mathbf{z}_l = \mathbf{a}_{l-1} sign(\mathbf{H}_l)$
9 | | | | Compute activations $\mathbf{a}_l = sign(\mathbf{z}_l)$
10 | | | | Compute local output $\hat{\mathbf{y}}_l = \mathbf{a}_l \mathbf{P}_l$
11 | | | | **foreach** $\mu = 1, \ldots, bs$ **do**
12 | | | | | Compute local 01-Loss $\ell_l^\mu = \mathcal{L}_{0/1}(\hat{\mathbf{y}}_l^\mu, y^\mu)$
13 | | | | | Compute $\tau_l^\mu = \hat{\mathbf{y}}_{l\,(c)}^\mu - \hat{\mathbf{y}}_{l\,(c-1)}^\mu$
14 | | | | | **if** $\ell_l^\mu = 1$ **or** $\tau_l^\mu < rK_l$ **then**
15 | | | | | | Add $\mu$ to set of pattern indexes to update $\mathcal{M}_l \leftarrow \mathcal{M}_l \cup \{\mu\}$
16 | | | | | **end**
17 | | | | **end**
18 | | | | *LayerUpdate*$(\mathbf{a}_{l-1}, \mathbf{y}, \mathbf{z}_l, \mathcal{M}_l, l, \gamma, p_r))$
19 | | | **end**
20 | | **end**
21 | | Rescale probability $p_r = p_r\sqrt{E^e}$
22 | **end**

$\{\pm 1\}$ (see line 2). Second, for each mini-batch $(\mathbf{x}, \mathbf{y}) \subseteq \mathbf{X}$, with $|\mathbf{x}| = |\mathbf{y}| = bs$, where $bs$ is the mini-batch size, the input patterns $\mathbf{x}$ are binarized into $\mathbf{a}_0$ using the median value as a threshold[1], resulting in $\mathbf{a}_{0,i}^\mu \in \{\pm 1\}$, $\mu \in \{1, \ldots, bs\}$, $i \in \{1, \ldots, K_0\}$ (see line 5). Third, the set of pattern indexes to update $\mathcal{M}_l$ is defined and initialized as the empty set $\emptyset$ (see line 7). At this point, the forward pass begins. For each layer $l \in \{1, \ldots, L\}$, the algorithm computes the pre-activations $\mathbf{z}_l = \mathbf{a}_{l-1} sign(\mathbf{H}_l)$, the activations $\mathbf{a}_l = sign(\mathbf{z}_l)$, and the local output $\hat{\mathbf{y}}_l = \mathbf{a}_l \mathbf{P}_l$ (see lines 8–10). It then evaluates, for each pattern index $\mu \in \{1, \ldots, bs\}$, whether the $l$th layer correctly classifies the binary input pattern $\mathbf{a}_0^\mu$ by computing the *01-Loss* $\ell_l^\mu = \mathcal{L}_{0/1}(\hat{\mathbf{y}}_l^\mu, y^\mu)$ defined as follows:

$$\mathcal{L}_{0/1}\left(\hat{\mathbf{y}^\mu}, y^\mu\right) = \begin{cases} 0 & \text{if } \arg\max\left(\hat{\mathbf{y}^\mu}\right) = y^\mu \\ 1 & \text{if } \arg\max\left(\hat{\mathbf{y}^\mu}\right) \neq y^\mu \end{cases}, \tag{1}$$

where $\hat{\mathbf{y}^\mu}$ is the local output of layer $l$ and $y^\mu$ is the true label, namely, the desired output. In addition to the computation of the *01-Loss* $\ell_l^\mu$, a stronger correctness constraint can be enforced by introducing a robustness parameter $r$. This user-specified hyperparameter is compared to the difference between the first and the second highest local output values $\tau_l^\mu = \hat{\mathbf{y}}_{l\,(c)}^\mu - \hat{\mathbf{y}}_{l\,(c-1)}^\mu$, which can be interpreted as the confidence of the $l$th layer in classifying the considered pattern $\mathbf{a}_0^\mu$. If the $\mu$th pattern is correctly classified, i.e. $\ell_l^\mu = 0$, with a confidence over the robustness threshold, i.e. $\tau_l^\mu \geqslant rK_l$ by the $l$th layer (notice that $r$ is scaled by the layer size $K_l$), no weight update is carried out. Otherwise, the index $\mu$ is added to the set of pattern indexes to update $\mathcal{M}_l$ of layer $l$ (see line 15). The robustness parameter has been shown to bias NNs toward flat regions in the loss landscape [41–44], and in appendix A.1 we verify that it benefits generalization. Once the mini-batch has been entirely processed, the algorithm proceeds at the layer-level by updating each layer $l$ simultaneously using its set $\mathcal{M}_l$, as described in the next section.

The novelty of the proposed architecture lies in the introduction of fixed and random binary local classifiers. Specifically, they generate binary, local, and randomized error signals, enabling the binary training of arbitrarily deep BMLPs. Moreover, this design supports the independent training of each layer and allows parallelization of the training procedure.

[1] Although alternative binarization methods can be considered, it is also feasible to adapt the BMLP to handle directly integer and floating-point inputs. This can be accomplished by adding a fixed and random expansion layer with weights in $\{\pm 1\}$ and a sign activation function as the input layer of the BMLP.

**Algorithm 2.** Proposed layer-level BMLP training algorithm.

**Variables:** $\boldsymbol{\rho}_l^{y^\mu}$ : random classifier weights in layer $l$ associated to the true label $y^\mu$ of the $\mu$th pattern

1 **def** *LayerUpdate*($\mathbf{a}_{l-1}, \mathbf{y}, \mathbf{z}_l, \mathcal{M}_l, l, \gamma, p_r$)**:**
2 $\quad \mathcal{G} \leftarrow$ (split $K_l$ into $G$ groups of size $\gamma$)
3 $\quad$ Initialize set of perceptron indexes to update $\mathcal{U}_l \leftarrow \emptyset$
4 $\quad$ **foreach** $\mu$ in $\mathcal{M}_l$ **do**
5 $\quad\quad$ Compute perceptron local stabilities $\boldsymbol{\delta}_l^\mu = \mathbf{z}_l^\mu \boldsymbol{\rho}_l^{y^\mu}$
6 $\quad\quad$ **foreach** $g$ in $\mathcal{G}$ **do**
7 $\quad\quad\quad$ Find *easiest* perceptron to fix $k_{l,g}^{\mu\star} = \arg\max_{k \in g}(\boldsymbol{\delta}_l^\mu : \boldsymbol{\delta}_l^\mu < 0)$
8 $\quad\quad\quad$ Add $(\mu, k_{l,g}^{\mu\star})$ to set of perceptron indexes to update $\mathcal{U}_l \leftarrow \mathcal{U}_l \cup \{(\mu, k_{l,g}^{\mu\star})\}$
9 $\quad\quad$ **end**
10 $\quad$ **end**
11 $\quad$ *PerceptronUpdate*($\mathbf{a}_{l-1}, \mathbf{y}, \mathcal{U}_l, l, p_r$)

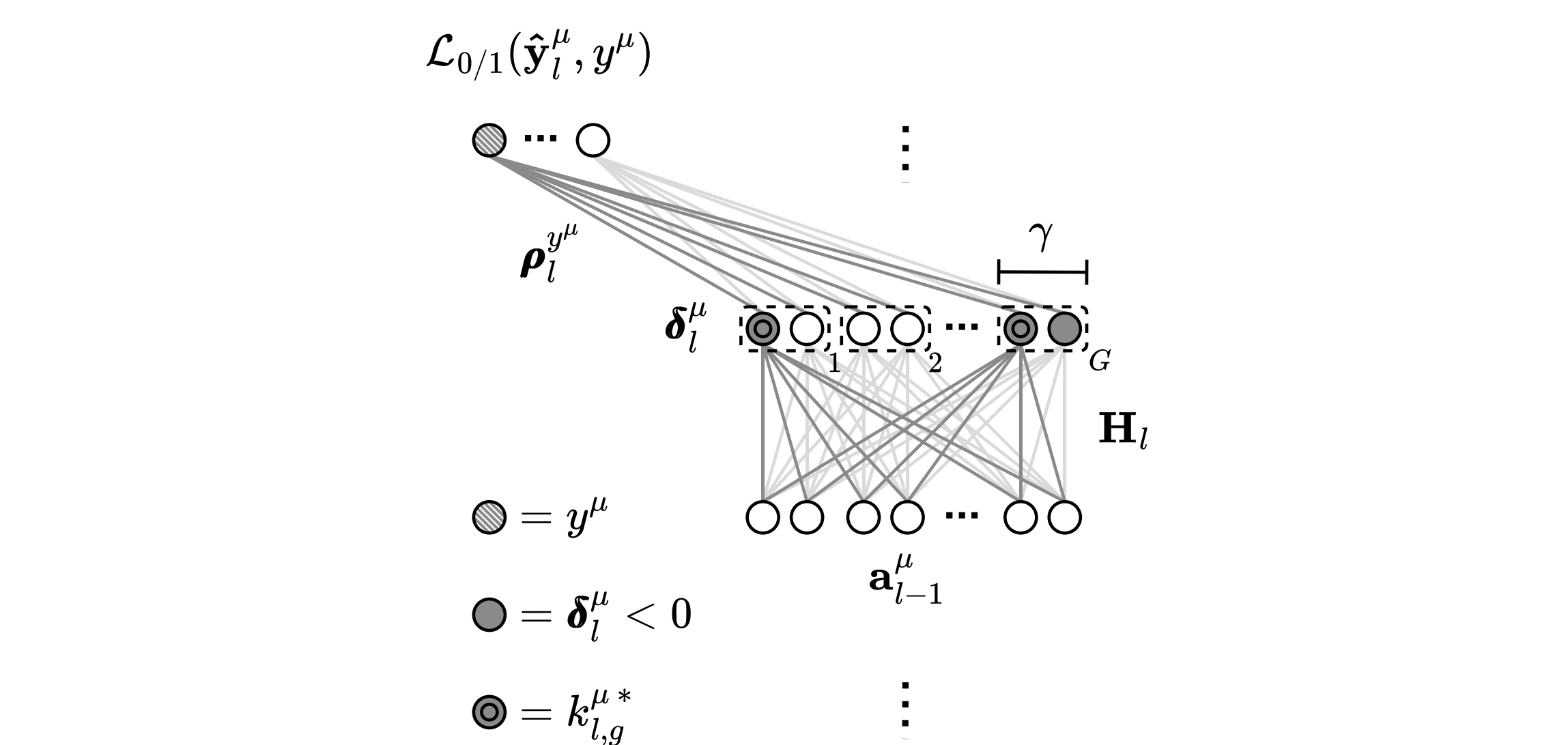

**Figure 2.** Proposed fully binary and gradient-free learning algorithm operating at the layer level. $\mathbf{H}_l$ are the hidden integer metaplastic weights, $\mathbf{a}_{l-1}$ are the activations of the previous layer, $G$ is the number of subgroups, $\gamma$ is the group size, $y^\mu$ is the true label of $\mu$th pattern, $\boldsymbol{\rho}_l^{y^\mu}$ are the random classifier weights associated to the true label, $\boldsymbol{\delta}_l^\mu$ are the local stabilities, and $k_{l,g}^{\mu\star}$ are the *easiest* perceptrons to update within each sub-group $g$.

### 3.3. Layer-level

If the set of pattern indexes to update $\mathcal{M}_l \neq \emptyset$ in a given layer $l$, i.e. $\ell_l^\mu = 1$ or $\tau_l^\mu < rK_l$ for at least one binary input pattern $\mathbf{a}_0^\mu$, with $\mu \in \{1, \ldots, bs\}$, as described in the previous section, the layer-level algorithm is executed. In particular, as detailed in algorithm 2 and illustrated in figure 2, it works as follows. First, the $K_l$ perceptrons are divided into $G$ subgroups of size $\gamma$, where $\gamma$ is a user-specified hyperparameter and $G = \frac{K_l}{\gamma}$ (see line 2). The role of $\gamma$ will be analyzed extensively in section 4.3. Second, the set of perceptron indexes to update $\mathcal{U}_l$ is defined and initialized as the empty set $\emptyset$ (see line 3). Third, each perceptron $k \in \{1, \ldots, K_l\}$ is examined to identify those contributing to the error of its associated random classifier $\mathbf{P}_l$. Specifically, the perceptrons contributing to the error are characterized by a negative product $\boldsymbol{\delta}_l^\mu$ of their pre-activations $\mathbf{z}_{l,k}^\mu$ and the random classifier weights associated with the true label $\boldsymbol{\rho}_l^{y^\mu}$. In other words, for each pattern $\mu \in \mathcal{M}_l$, the perceptrons for which the local stabilities $\boldsymbol{\delta}_l^\mu = \mathbf{z}_l^\mu \boldsymbol{\rho}_l^{y^\mu} < 0$ are considered (see line 5). Lastly, within each sub-group $g \in \{1, \ldots, G\}$, only the *easiest* perceptron to fix is selected (i.e. the one with the negative stability closest to 0), whose index is given by $k_{l,g}^{\mu\star} \in \arg\max_{k \in g}(\boldsymbol{\delta}_l^\mu : \boldsymbol{\delta}_l^\mu < 0)$ (see line 7). This heuristic targets the neuron whose activation is closest to a correct contribution, requiring the smallest adjustment to its underlying integer hidden weight to flip its value.

The selected perceptron indexes, which amount to at most $G$ (i.e. when there is at least one perceptron contributing to the error per group), along with their corresponding pattern index $\mu$, are added to the set $\mathcal{U}_l$ (see line 8). Once every $\mu \in \mathcal{M}_l$ has been processed, the set $\mathcal{U}_l$ contains at most $G \times bs$ tuples, each of which contains the pattern index $\mu$ and the perceptron index $k_{l,g}^{\mu\star}$ (e.g. $\mathcal{U}_l = \{(\mu_1, k_{l,1}^{\mu_1\star}), \ldots, (\mu_1, k_{l,G}^{\mu_1\star}), \ldots, (\mu_{bs}, k_{l,1}^{\mu_{bs}\star}), \ldots, (\mu_{bs}, k_{l,G}^{\mu_{bs}\star})\}$). These tuples are then used by the perceptron-level algorithm to perform the updates, as described in the next section.

**Algorithm 3.** Perceptron-level BMLP training algorithm [31].

**Variables:** $\mathbf{h}_l^k$: hidden metaplastic weights in layer $l$ associated to perceptron $k$

1 **def** *PerceptronUpdate*($\mathbf{a}_{l-1}, \mathbf{y}, \mathcal{U}_l, l, p_r$)**:**
  // `Clipped Perceptron`
2   **foreach** $(\mu, k_{l,g}^{\mu\,\star})$ in $\mathcal{U}_l$ **do**
3     Update $\mathbf{h}_l^{k_{l,g}^{\mu\,\star}} \leftarrow \mathbf{h}_l^{k_{l,g}^{\mu\,\star}} + 2\,\mathbf{a}_{l-1}^{\mu}\rho_l^{y^{\mu},k_{l,g}^{\mu\,\star}}$
4   **end**
  // `Reinforcement`
5   **foreach** $k = 1, \ldots, K_l$ **do**
6     **foreach** $h$ in $\mathbf{h}_l^k$ **do**
7       Extract $p \leftarrow \mathrm{Uniform}(0,1)$
8       **if** $p < p_r\sqrt{\frac{2}{\pi K_l}}$ **then**
9         $h \leftarrow h + 2sign(h)$
10       **end**
11     **end**
12   **end**

**Table 3.** Computational costs of the proposed solution in terms of number of operations for each layer $l = \{1, \ldots, L\}$ and for each input pattern $\mathbf{a}_0^{\mu}$. $K_l$ and $K_{l-1}$ are the sizes of layer $l-1$ and layer $l$, respectively. $c$ is the number of classes, and $\gamma$ is the group size.

| | Number of operations | |
|---|---|---|
| Operation | Forward | Backward |
| XNOR | $K_l(K_{l-1}+c)$ | $K_l\left(1+\frac{K_{l-1}}{\gamma}\right)$ |
| Popcount | $K_l+c$ | — |
| Increment/decrement | — | $2\frac{K_lK_{l-1}}{\gamma}$ |

The novelty of the proposed learning algorithm lies in its strategy of grouping neurons and updating only the *easiest* one. This approach introduces a sparse update scheme, which, as shown in section 4.3, acts as an effective regularizer that enhances generalization.

### 3.4. Perceptron level

Once the perceptrons to update $\mathcal{U}_l$ have been selected for each layer $l$, as described in the previous section, the proposed algorithm proceeds at the perceptron level. Specifically, as detailed in algorithm 3, it relies on the two steps of the CP+R rule [31]. First, for each tuple $(\mu, k_{l,g}^{\mu\star}) \in \mathcal{U}_l$, it updates the hidden metaplastic weights $\mathbf{h}_l^{k_{l,g}^{\mu\,\star}}$ (i.e. those associated with the previously selected perceptron $k_{l,g}^{\mu\,\star}$) by using the Clipped Perceptron (CP) rule, i.e. $\mathbf{h}_l^{k_{l,g}^{\mu\,\star}} \leftarrow \mathbf{h}_l^{k_{l,g}^{\mu\,\star}} + 2\,\mathbf{a}_{l-1}^{\mu}\rho_l^{y^{\mu},k_{l,g}^{\mu\,\star}}$ (see lines 2–4). Second, it computes the Reinforcement (R) rule of each metaplastic hidden weight $h \in \mathbf{H}_l$ according to the reinforcement probability $p_r$, i.e. $h \leftarrow h + 2sign(h)$ (see lines 5–12). At the end of each epoch $e$, the reinforcement probability $p_r$ is rescaled by a factor $\sqrt{E^e}$, where $E^e$ is the fraction of training errors made by the BMLP during the current epoch $e$.

### 3.5. Computational costs and memory footprint

The proposed fully binary and gradient-free learning algorithm offers two key advantages with respect to (w.r.t.) full-precision SGD. Firstly, from a computational perspective, it enables the exploitation of efficient binary operations even during the training phase, thereby drastically reducing computational complexity and execution time. Specifically, it is capable of training a BMLP relying solely on XNOR, Popcount, and increment/decrement operations. Table 3 illustrates the number of operations required to process a single input pattern $\mathbf{a}_0^{\mu}$ for each layer $l$, both in the forward and the backward passes. Second, from a memory perspective, it reduces the memory footprint of both the training procedure and the final model. Specifically, as summarized in table 4, the activations $\mathbf{a}_l$ and the binary weights $\mathbf{w}_l$ require only 1-bit values, while the integer-valued hidden metaplastic weights $\mathbf{h}_l$ require 8-bit values. It is worth noting that, similarly to all the other BNN-related works present in the literature [8, 10], the pre-activations $\mathbf{z}_l$ require $\lceil \log_2(2K_{l-1}) \rceil$ bits to be represented. Nonetheless, the values of $\mathbf{z}_l$ can be bounded by using a bit-width equal to $\min_{l\in\{1,\ldots,L\}}(8, \lceil \log_2(2K_{l-1}) \rceil)$, without compromising the final accuracy.

We emphasize that the proposed algorithm requires only metaplastic hidden levels to develop inertia, representing each synapse's confidence in its visible sign. Consequently, the algorithm can be implemented using bit-shift operations instead of increment/decrement operations, albeit at the cost of increasing the bit-width of the integer-valued hidden metaplastic weights $\mathbf{h}_l$ to ensure a sufficient number of levels. The exploration of this aspect is left for future work.

**Table 4.** Memory footprint of the proposed solution in terms of number of bits needed for representing each value. $\mathbf{a}_l$ are the activations, and $\mathbf{h}_l$ and $\mathbf{w}_l$ are the integer-valued integer metaplastic weights and the binary weights, respectively. The hidden weights $\mathbf{h}_l$ are used only during the backward pass, while the binary weights $\mathbf{w}_l$ are used only during the forward pass.

| | Number of bits | |
|---|---|---|
| Variable | Forward | Backward |
| $\mathbf{a}_l$ | 1 | 1 |
| $\mathbf{h}_l$ | — | 8 |
| $\mathbf{w}_l$ | 1 | — |

## 4. Experimental results

In this section, we evaluate our proposed fully binary and gradient-free training algorithm for multi-layer BNNs on a range of benchmark datasets. The objective is twofold:

1. Demonstrate that our method surpasses the performance of the single-layer fully binary SotA algorithm [10].
2. Compare our solution to MLPs trained with full-precision SGD under the same total memory and computational demands, where all weights, activations, and gradients are represented using floating-point values.

The Python code of the experiments performed in this paper is made available to the scientific community as a public repository[2].

Section 4.1 describes the benchmark datasets used in the experimental campaign. In section 4.2, the proposed algorithm is compared with the SotA solution [10] and full-precision SGD, under both memory and computational demand constraints. Section 4.3 analyzes the role of the group size $\gamma$ in the training procedure, while section 4.4 presents an ablation study on three enhanced versions of the SotA algorithm [10]. Lastly, in section 4.5, the effectiveness of the proposed solution is evaluated on deep multi-layer BNNs.

### 4.1. Datasets

*4.1.1. Random Prototypes*

We generate a synthetic dataset of $K_0$-dimensional binary vectors sampled from $\{\pm 1\}$ by first creating one random prototype vector per class. Each prototype is then used to generate class samples by flipping each of its entries with a user-defined probability $p$, which controls the difficulty of the task. This process clusters samples around their respective prototypes in a sign-flipping analog of Hamming distance, enabling both optimization and generalization. To ensure uniqueness, we discard any repeated vectors and continue sampling until the desired number of unique samples is reached. In our experiments, we set $K_0 = 1000$ and $p = 0.44$. The final dataset comprises 10 000 training samples and 2000 test samples across 10 different classes.

*4.1.2. FashionMNIST*

FashionMNIST [45] is a dataset of 28 × 28 grayscale images of Zalando articles, designed as a more challenging alternative to the MNIST dataset [46]. It consists of 50 000 training images and 10 000 test images across 10 different classes.

*4.1.3. CIFAR-10*

CIFAR10 [47] consists of 32 × 32 color images across 10 object classes, with 40 000 training samples and 10 000 test samples. Specifically, we consider features extracted from a pre-trained AlexNet [48], a CNN composed of five convolutional layers with ReLU activation functions [49] and three max pooling layers, on which we train a BMLP classifier.

*4.1.4. Imagenette*

Imagenette [50] is a simplified 10-class subset of the larger ImageNet dataset [51]. It comprises 32 × 32 color images, with 10 000 training samples and 3394 test samples across 10 classes. We use features extracted from a pre-trained AlexNet [48], on which we train a BMLP classifier.

[2] https://github.com/AI-Tech-Research-Lab/BNN.

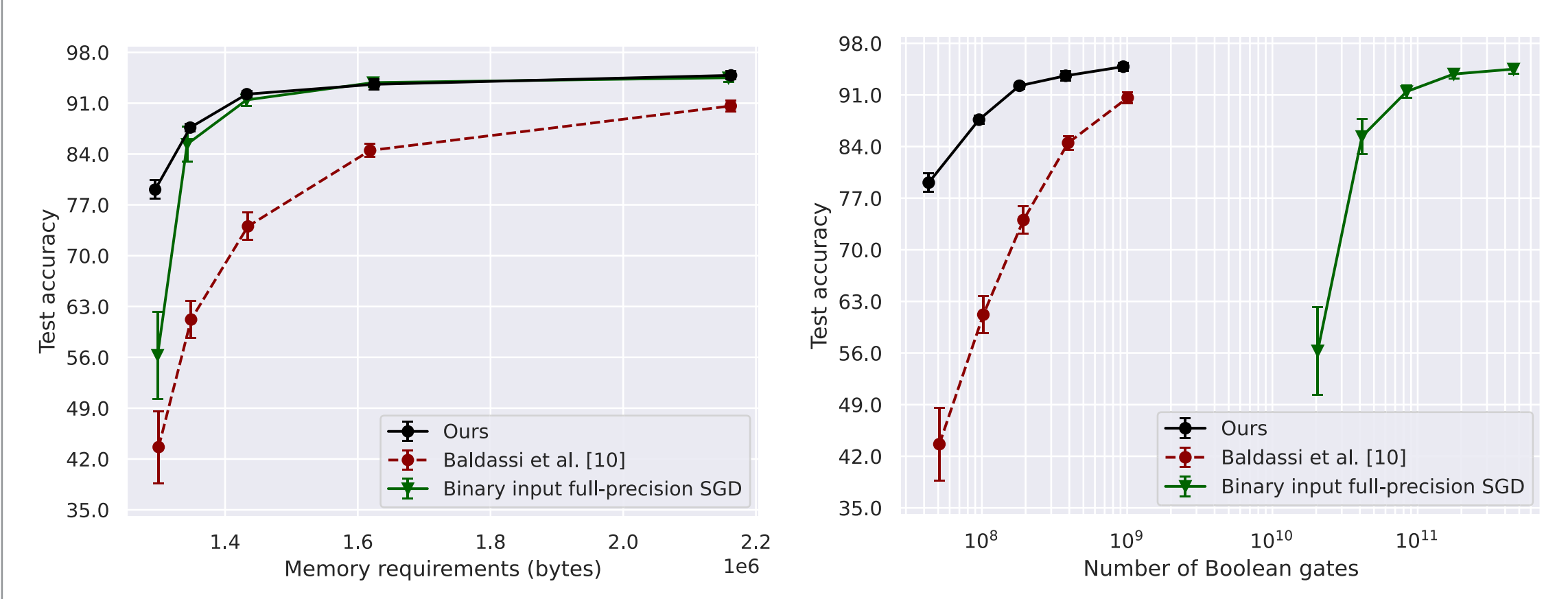

(a) Test accuracy w.r.t. memory requirements (b) Test accuracy w.r.t. number of Boolean gates

**Figure 3.** Test accuracy (averaged over 10 realizations of the initial conditions) as a function of total memory requirements and Boolean gate count on the Random Prototypes dataset. The results compare our multi-layer fully binary and gradient-free algorithm with both the fully binary single-layer SotA solution [10] and full-precision SGD.

### 4.2. Comparison with the SotA algorithm [10] and full-precision SGD

In this section, we evaluate our proposed algorithm in the challenging scenario of low memory and computational budgets, considering both the total memory footprint and the total number of Boolean gates used during training, as detailed later in this section. Specifically, we compare our method with the fully binary single-layer SotA algorithm [10][3]. We emphasize that, while both algorithms avoid floating-point backpropagation, the one introduced in [10] is limited to training a single hidden layer and relies on a custom architecture with a sparse grouping layer, where 0 is introduced as a third weight representation alongside $\pm 1$, effectively making the resulting NN *ternary*. The proposed solution is used to train two hidden layer BNNs with the following set of hyperparameters $\mathcal{H}$: batch size $bs = 100$, number of epochs $e = 50$, probability of reinforcement $p_r = 0.5$, and robustness $r = 0.25$. To demonstrate the memory and computational advantages of our algorithm, we also trained two hidden layer floating-point MLPs using full-precision SGD, with the following set of hyperparameters $\mathcal{H}$: batch size $bs = 100$, number of epochs $e = 50$, and learning rate $\eta = 0.01$. Specifically, we train floating-point MLPs on both full-precision and binarized versions of the input datasets under the same memory requirement, reducing the size of the training set accordingly when floating-point inputs are considered. To ensure a fair comparison of accuracy, we selected model architectures that equalize total memory usage across methods. This isolates the expressive power of the training rules themselves, regardless of differences in memory budgets. Notably, we compare our method against full-precision SGD, which serves as a practical upper bound on the accuracy of any BNN trained using QAT [8, 16, 17, 25, 26]. For completeness, results using Adam optimizer [52] are provided in A.3.

Figure 3(a) shows the test accuracy, averaged over 10 different runs, as a function of total memory requirements (including both model and dataset) on the Random Prototypes dataset. Similarly, figure 4(top panels) presents analogous results for the FashionMNIST, CIFAR-10, and Imagenette datasets. Our proposed solution consistently outperforms the SotA algorithm [10], achieving test accuracy improvements of up to +35.47%, +24.80%, +31.82% and +25.10% on the Random Prototypes, FashionMNIST, CIFAR-10, and Imagenette datasets, respectively, at the lowest memory requirements. Furthermore, the results obtained using our proposed solution exhibit lower variance compared to those obtained by [10], indicating a more stable learning procedure.

Compared to full-precision SGD on the same binarized datasets, our approach achieves comparable test accuracy across various memory requirements, with improvements at the lowest ones up to +22.82%, +0.65%, +10.25% and +9.89% on the Random Prototypes, FashionMNIST, CIFAR-10 and Imagenette datasets, respectively. In comparison to SGD on the floating-point version of the dataset, we observe test accuracy gains of up to +29.18%, +21.55% and +35.30% on the FashionMNIST, CIFAR-10 and Imagenette datasets, respectively (the Random Prototypes dataset is binary by definition). This result highlights another interesting finding, i.e. at the same fixed memory demand, training full-precision SGD on a larger set of binarized samples is more effective than training it on a smaller set of full-precision ones.

[3] Since [10] does not provide open-source code, we replicated it as faithfully as possible based on the published description.

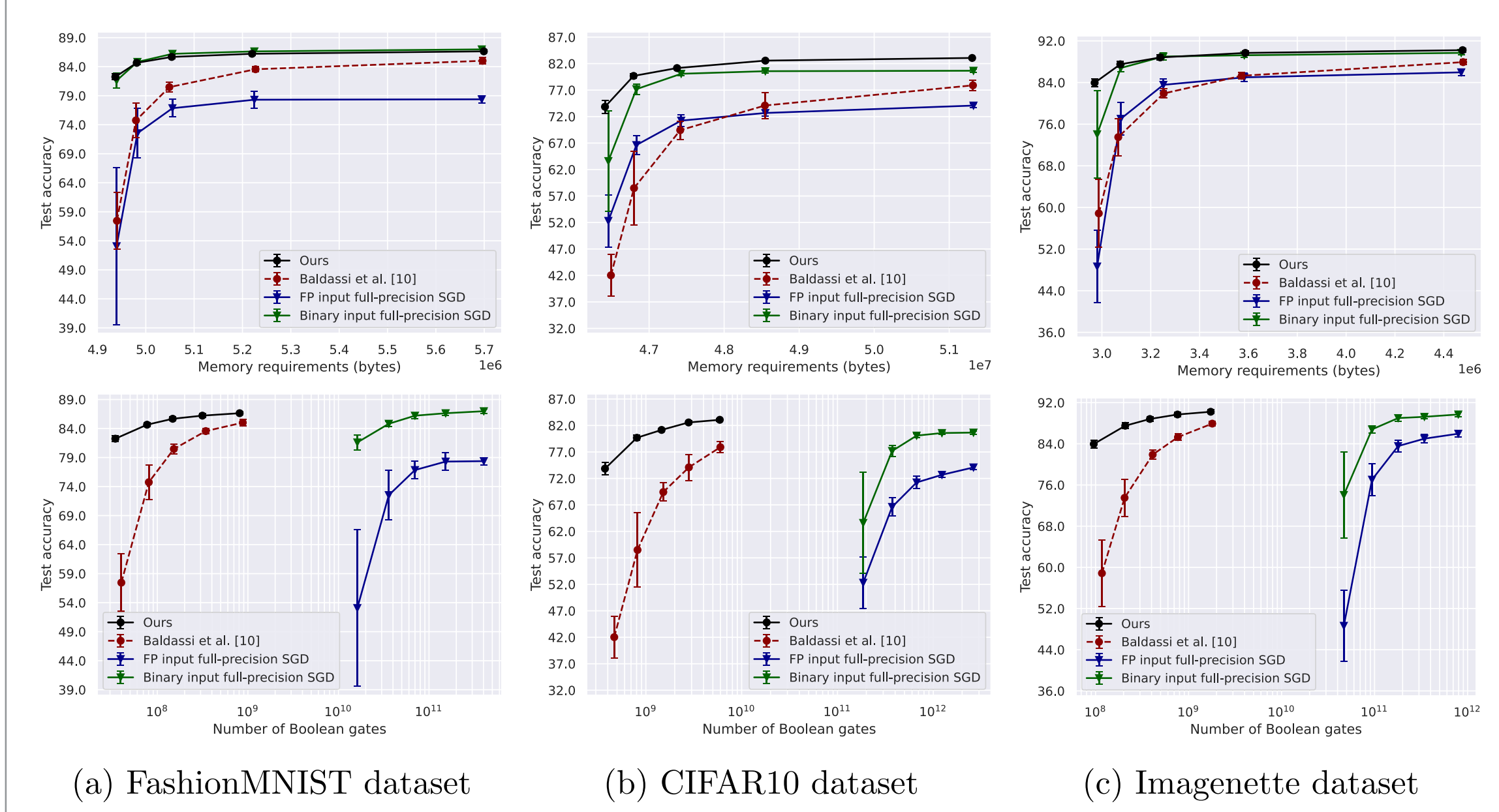

(a) FashionMNIST dataset (b) CIFAR10 dataset (c) Imagenette dataset

**Figure 4.** The results compare our multi-layer fully binary and gradient-free algorithm with both the fully binary single-layer SotA solution [10] and full-precision SGD. (Top panels) Test accuracy (averaged over 10 realizations of the initial conditions) as a function of total memory requirements. (Bottom panels) Test accuracy (averaged over 10 realizations of the initial conditions) as a function of the Boolean gate count.

These results become even more significant when considering the drastic reduction in computational complexity and the efficiency of the operations used in our fully binary and gradient-free training algorithm (i.e. XNOR, Popcount, and increment/decrement). To quantify the computational cost at a hardware level, we estimate the total number of Boolean gates required for a single training step (i.e. the forward and backward passes for one batch) and compare it to the number of equivalent Boolean gates needed for floating-point additions and multiplications in full-precision SGD training. This metric captures the inherent complexity of arithmetic operations, enabling a more fundamental comparison of computational cost than simply counting high-level operations. It also aligns with similar metrics explored in other works, which use the number of binary operations to account for the efficiency of logic- or binary-based NNs [17, 19, 53, 54]. More specifically, IEEE-754 single-precision (32-bit) floating-point additions and multiplications [55] are estimated to require on the order of $10^4$ equivalent Boolean gates [56–59], whereas $N$-bit Popcount and increment/decrement operations on $N$-bit integers require at most $\mathcal{O}(10N)$ Boolean gates [60–64]. Figures 3(b) and 4(bottom panels) show that our algorithm requires two to three orders of magnitude fewer Boolean gates during training. Furthermore, our proposed solution achieves an optimal tradeoff between accuracy and the number of Boolean gates, even when compared to [10].

As demonstrated in this section, our proposed solution consistently outperforms the fully binary single-layer SotA algorithm [10] in test accuracy across all tested datasets. Moreover, it crucially matches or even surpasses full-precision SGD under the same memory and computational requirements. This can be attributed to the fact that using binarized data allows for the processing of significantly more information (by a factor of 32) compared to single-precision floating-point inputs, which is essential for effectively training DL models that require large amounts of training samples.

### 4.3. The role of the group size $\gamma$

A crucial element in enhancing the generalization of our proposed fully binary learning algorithm is partitioning neurons in each hidden layer into groups of size $\gamma$, as described in section 3.3. In this section, we investigate the impact of varying the group size $\gamma$ on the test accuracy results of a single hidden layer BMLP using our proposed learning algorithm on the FashionMNIST, CIFAR10, and Imagenette datasets. The hyperparameters $\mathcal{H}$ are set as follows: batch size $bs = 100$, number of epochs $e = 50$, probability of reinforcement $p_r = 0.5$, and robustness $r = 0.25$. Figure 5 shows the test accuracy, averaged over 10 realizations of the initial conditions, as a function of the group size $\gamma$.

Interestingly, an optimal range $\Gamma^*$ emerges independently of the hidden layer size $K_l$. This implies that the optimal number of groups, and consequently the number of updated perceptrons at each step, is proportional to the hidden layer size. In other words, there exists an optimal value of the ratio $\frac{\gamma^*}{K_l}$, or, equivalently, an optimal fraction of updated neurons w.r.t. the hidden layer size $\frac{\text{number of updated neurons}}{\text{hidden layer size}}$, which is independent of the hidden layer size itself. Furthermore, as can be seen in figure 5, the closer $\gamma$ is to the optimal value $\gamma^*$, the lower the variance in test accuracy across different initializations. This highlights the crucial role of the group size $\gamma$ not only in the generalization of the final BMLP but also in ensuring the stability of the learning procedure.

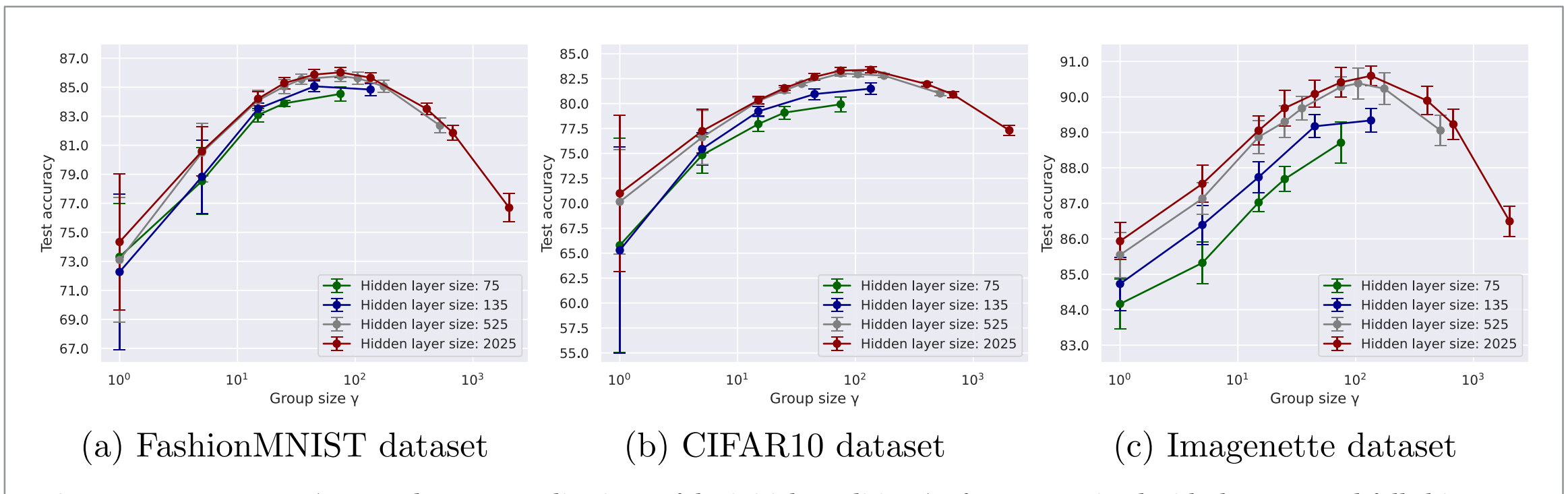

**Figure 5.** Test accuracy (averaged over 10 realizations of the initial conditions) of BMLPs trained with the proposed fully binary algorithm as a function of the group size $\gamma$ for different hidden layer sizes.

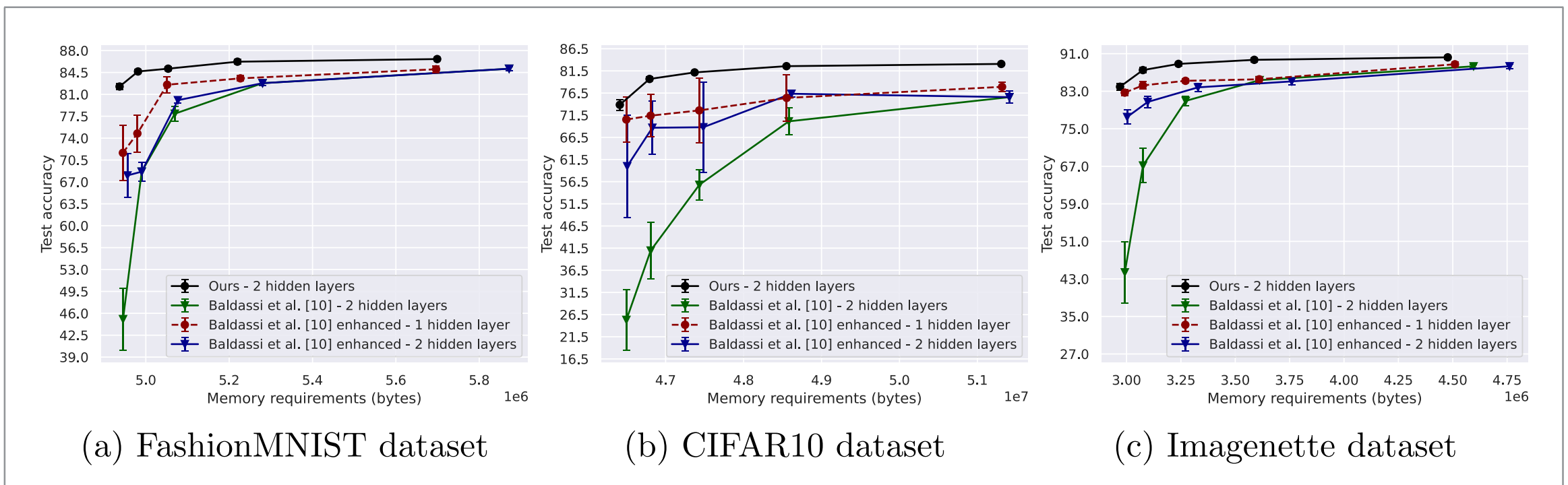

**Figure 6.** Test accuracy (averaged over 10 realizations of the initial conditions) as a function of total memory requirements. The results compare our multi-layer fully binary and gradient-free algorithm with three enhanced variants of the fully binary single-layer SotA solution [10].

Based on empirical findings, the optimal range for $\gamma$ lies approximately within $\Gamma^* \approx [75, 105]$ for the considered datasets and architectures. Accordingly, fine-tuning $\gamma$ should consider values close to this empirically determined range. In practice, the optimal value $\gamma^* \in \Gamma^*$, for a generic layer of size $K_l$, is chosen as $\gamma^* = \arg\min_{\gamma \in \{\gamma | \gamma \text{divides } K_l\}} \min_{\tilde{\gamma} \in \Gamma^*} |\gamma - \tilde{\gamma}|$, i.e. selecting the divisor of the layer size closest to any value in the optimal range $\Gamma^*$. This value serves as a principled starting point for subsequent fine-tuning. For instance, in a model characterized by a hidden layer of size 525, $\gamma$ is set to 105, resulting in the layer being divided into 5 equal-sized groups. Similarly, in a model characterized by a hidden layer of size 2025, $\gamma$ is set to 81, leading to a layer divided into 25 equal-sized groups. These values of $\gamma$ imply that our algorithm achieves the best performances for very *sparse* updates: only approximately 0.9% and 1.3% of the perceptrons in each layer are updated at every step, respectively. This is a very small fraction compared to standard SGD training, which updates 100% of the units in the NN at each step. We hypothesize that the grouping parameter $\gamma$ introduces a regularization effect. Specifically, it prevents a small subset of neurons from dominating the learning process by encouraging updates to be more evenly distributed across the layer, thereby promoting balanced learning throughout the entire BMLP.

It is relevant to note that, due to the permutation symmetry of the hidden layers, neurons in each group can be selected once at random and remain fixed throughout the learning process. We observed, however, that randomly reshuffling group membership at each update step does not impact the effectiveness of the proposed method. In contrast, selecting a random fraction of the easiest perceptrons to update, without dividing them into subgroups, negatively impacts the generalization of the proposed algorithm. A deeper investigation of these aspects, including a rigorous analysis of the regularization effect and its influence on the learning dynamics, remains an open problem for future research.

### 4.4. Ablation study: three enhanced variants of the SotA algorithm [10]

To further confirm and evaluate the effectiveness of our proposed algorithm, we conduct a broad ablation study comparing our method to three enhanced versions of the SotA algorithm [10]. First, we perform a search (detailed in A.2) to optimize the dimensions of the custom classification layer used in the SotA algorithm [10] and apply the optimal layer sizes to each considered NN architecture. Second, we extend the SotA algorithm [10] to support multi-layer BNNs by incorporating local classifiers. Third, we combine both the classification layer dimensions optimization and the multi-layer extension. Specifically, we train these three variants on one and two hidden layer BMLPs using the following set of hyper-parameters: batch size $bs = 100$, number of epochs $e = 50$, probability of reinforcement $p_r = 0.5$, and robustness $r = 0.25$.

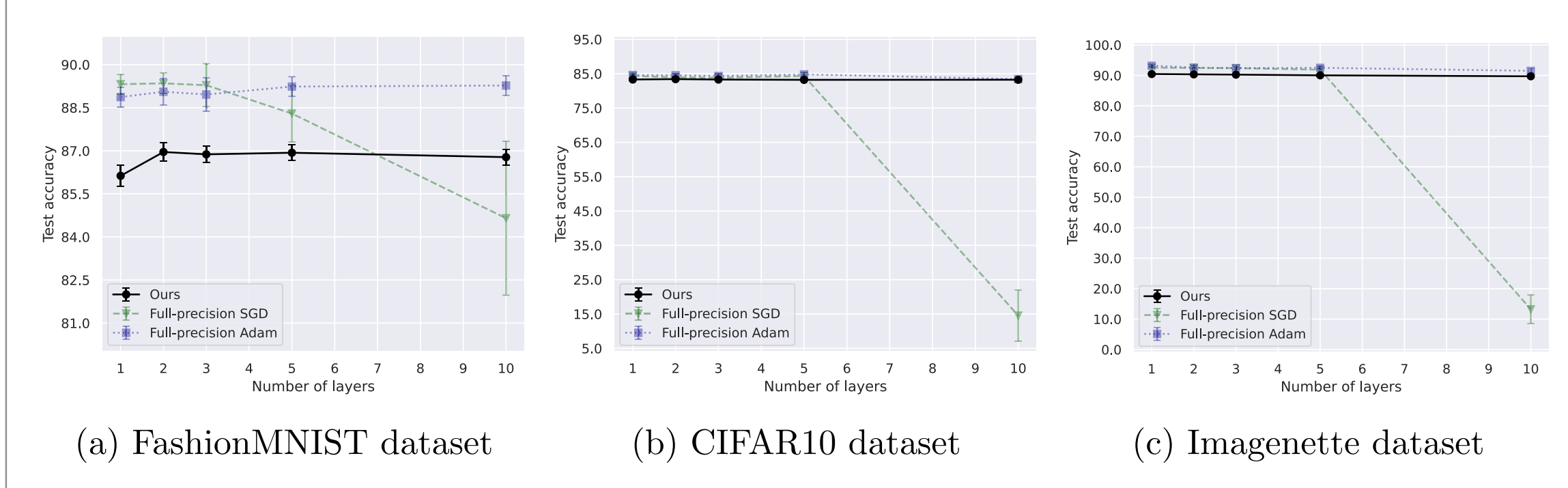

(a) FashionMNIST dataset (b) CIFAR10 dataset (c) Imagenette dataset

**Figure 7.** Test accuracy (averaged over 10 realizations of the initial conditions) as a function of the number of hidden layers $L$. The results compare our multi-layer fully binary and gradient-free algorithm with both full-precision SGD and Adam baselines.

Figure 6 shows the test accuracy, averaged over 10 realizations of the initial conditions, as a function of total memory requirements across the FashionMNIST, CIFAR-10, and Imagenette datasets, comparing our proposed algorithm to the three enhanced variants of [10]. All three versions successfully train BMLPs with both one and two hidden layers. Nonetheless, our proposed method consistently outperforms them across all datasets. While the explored enhancements mitigate some limitations of the SotA algorithm [10], our fully binary multi-layer algorithm still achieves a superior accuracy-memory trade-off.

### 4.5. Training deep multi-layer BNNs

To demonstrate that our proposed algorithm is capable of training deep BNNs with more than two hidden layers, we fix the hidden layer size $K_l$ to 1035 and train BMLPs with a number of layers $L \in \{1, 2, 3, 5, 10\}$ using the following hyperparameters: batch size $bs = 100$, number of epochs $e = 50$, reinforcement probability $p_r = 0.5$, and robustness $r_l = 0.25$. As a baseline, we train floating-point MLPs with the same number and size of layers using full-precision SGD and Adam, with the following hyperparameters: batch size $bs = 100$, number of epochs $e = 50$, and learning rates $\eta = 0.01$ for SGD and $\eta = 0.001$ for Adam, respectively. We emphasize that, differently from section 4.2, the experiments performed in this section do not directly compare our solution with full-precision SGD and Adam, since using the same number of parameters would result in significantly higher memory requirements and computational complexity for full-precision training.

Figure 7 shows the test accuracy, averaged over 10 realizations of the initial conditions, as a function of the number of hidden layers $L$, for the three training methods. Our proposed solution successfully trains BMLPs with an arbitrary number of layers, exhibiting test accuracy behavior consistent with that of MLPs trained using floating-point Adam, i.e. it saturates quickly for deeper architectures. In contrast, full-precision SGD fails to maintain stable and effective learning in deep BMLPs with 10 hidden layers. Notably, the ability to train deep BNNs is crucial, as it establishes a foundational methodology for integrating additional layers (e.g. convolutional ones) and extending the proposed solution to larger, state-of-the-art models.

Overall, the experimental results confirm that our proposed fully binary and gradient-free learning algorithm effectively handles both wide (high-parameter) and deep (many-layer) architectures, conserving most of the accuracy of floating-point MLPs trained with full-precision SGD and Adam, while dramatically reducing computational cost and memory usage.

## 5. Discussion and limitations of our proposed solution

Despite its promising results, our proposed solution has some limitations that suggest directions for future research. First, although significantly more efficient than floating-point representations, the current framework's reliance on integer-valued hidden weights prevents a purely 1-bit implementation. Achieving effective learning without this form of synaptic memory remains a substantial challenge, particularly due to catastrophic forgetting, as investigated in A.5. Second, this study focuses on fully-connected layers and MLPs. Extending these principles to other architectures, such as CNNs or recurrent NNs (RNNs), is a non-trivial but essential step toward broader applicability. More specifically, the proposed local learning rule is not directly compatible with the core properties of convolutional and recurrent layers, namely spatial locality and weight sharing. Extending our fully binary training paradigm to CNNs and RNNs would require the development of an end-to-end binary backpropagation rule capable of handling these structural priors. This remains a significant research challenge, which we identify as a key direction for future work, as discussed in section 6. Finally, the proposed solution is currently designed for classification tasks only. In particular, the binary nature of the output layer limits its ability to represent continuous values. Supporting regression tasks would require the introduction of encoding schemes such as thermometer encoding [64] or other structured binary representations of floating-point values at the output layer.

It is also important to contextualize our work w.r.t. conventional BNN architectures, such as VGG [2], MobileNet [65], ResNet [66], and Yolo [67]. As detailed in sections 2.1 and 2.4, most existing BNN models are trained using QAT, which relies on full-precision SGD during training. This prevents the backward pass from being binary. Furthermore, state-of-the-art BNN architectures typically incorporate components such as batch normalization, full-precision input/output layers, and skip connections. As a result, these models are not fully binary during inference either. In contrast, our proposed solution maintains binary computation throughout both the training and inference phases, enabling significant reductions in both computational cost and memory demand.

## 6. Conclusions and future research directions

In this paper, we introduced a fully binary and gradient-free learning algorithm based on binary forward and binary backward passes for training BMLPs, bridging the gap between hardware-friendly 1-bit arithmetic and biologically plausible randomized local learning rules. The proposed solution advances the state-of-the-art by enabling the training of multi-layer BNNs using only random local binary error signals and binary weight updates, entirely discarding the full-precision floating-point SGD-based backpropagation algorithm typically used for BNN training. In particular, our forward and backward steps rely solely on XNOR, Popcount, and increment/decrement operations. This design not only makes BNN training extremely compute- and memory-efficient, but also draws inspiration from neurobiology, leveraging randomized binary local errors and integer metaplastic hidden weights to mitigate catastrophic forgetting. Experimental evaluations on multi-class classification benchmarks show test accuracy improvements of up to +35.47% over the fully binary single-layer SotA algorithm [10]. Compared to full-precision SGD, our solution improves test accuracy by up to +41.31% under the same total memory demand, while also reducing computational cost by two to three orders of magnitude in terms of the total number of Boolean gates.

The application scenarios that can benefit from our solution are mainly three. The first is TinyML, which targets devices with strict resource constraints in terms of processing power, memory, and energy [68]. In this case, our proposed solution could enable tiny devices to train ML models locally without relying on external cloud infrastructures, allowing real-time processing, reducing energy consumption, and enhancing data privacy. Real-world applications include on-device keyword spotting [69, 70], real-time anomaly detection [71, 72], and lightweight medical monitoring tasks such as face mask detection [73, 74]. The second scenario is PPML, which aims to protect the confidentiality of sensitive data during the training and deployment of ML and DL solutions in a *as-a-service* manner [75]. Methods such as homomorphic encryption (HE) enable computations on encrypted data, but are notoriously expensive when involving floating-point multiplications and additions [75, 76]. By leveraging purely Boolean arithmetic, our approach holds promise for making HE-based privacy-preserving ML more feasible, potentially mitigating the significant overhead of encrypted NN training [77]. This could enable real-life applications involving sensitive financial or healthcare data to benefit from the advantages of secure computation within a practical timeframe [78, 79]. The third scenario is learning in neuromorphic hardware, such as spiking NNs (SNNs) [80], which are designed to mimic biological neural processes and operate efficiently on specialized hardware. Our algorithm shares conceptual similarities with learning in SNNs. These models process information through sparse, binary events (i.e. spikes), and their learning rules are often local and time-dependent, such as spike-timing-dependent plasticity [81]. Similarly, the use of local binary error signals and binary weight updates in our proposed solution could provide a blueprint for developing efficient and non-gradient-based on-chip learning algorithms for neuromorphic hardware. This is particularly promising for applications such as autonomous robotics or smart implants, where power efficiency and low latency are critical [82, 83]. These systems require rethinking traditional NN learning algorithms, as conventional gradient-based methods, such as backpropagation, are not easily implementable on-chip [84].

As discussed in section 5, several directions for future research remain. *(i)* While in this paper we focused on fully connected layers, extending the method to binary convolutional and graph NNs is a natural and crucial non-trivial next step for computer vision and graph learning tasks. *(ii)* While we investigated *local* binary error signals, an interesting direction is to explore whether the quality of the binary error signals can be improved by designing an end-to-end binary backward rule for BNN training, akin to the well-known end-to-end gradient backpropagation rule. *(iii)* The experimental results presented in this paper suggest a regularizing effect from partitioning neurons into subgroups of size $\gamma$. A deeper theoretical analysis could clarify its impact on the convergence of learning dynamics and generalization of our algorithm. *(iv)* Lastly, pure binary training, i.e. training without hidden metaplastic integer variables, will also be investigated. However, preliminary evidence suggests that catastrophic forgetting may become more severe in this extreme scenario, highlighting the need for additional and innovative strategies.

## Data availability statement

The data that support the findings of this study are openly available at the following URL/DOI: https://github.com/AI-Tech-Research-Lab/BNN.

## Acknowledgments

This paper is supported by Dhiria s.r.l. and by PNRR-PE-AI FAIR project funded by the NextGeneration EU program.

## Appendix A. Ablation studies

### A.1. The robustness hyperparameter *r*

In this section, we perform an ablation study on the robustness parameter $r$, as shown in figure A1. Notably, an optimal value for the robustness parameter emerges across all considered tasks, i.e. $r = 0.25$, allowing us to keep it fixed in all our experimental evaluations.

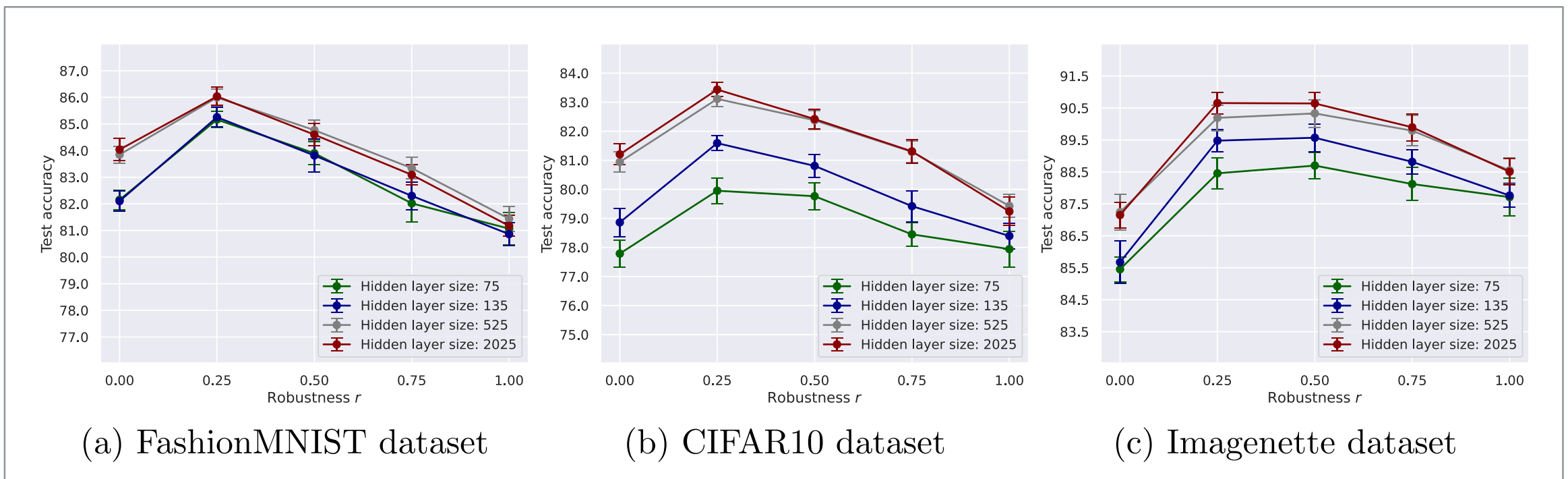

(a) FashionMNIST dataset (b) CIFAR10 dataset (c) Imagenette dataset

**Figure A1.** Test accuracy (averaged over 10 realizations of the initial conditions) of BMLPs trained with the proposed fully binary algorithm as a function of the robustness hyperparameter $r$ for different hidden layer sizes.

### A.2. Optimizing the classification layer dimensions in the SotA algorithm [10]

The algorithm proposed in [10] includes a hyperparameter that defines the dimension of the fixed (non-learnable) classification layer. However, its optimal value was not explored in the original paper. To ensure a fair comparison with our algorithm, we perform a comprehensive hyperparameter search and use the optimal value in the experiments of section 4.4. The results of this search are presented in figure A2.

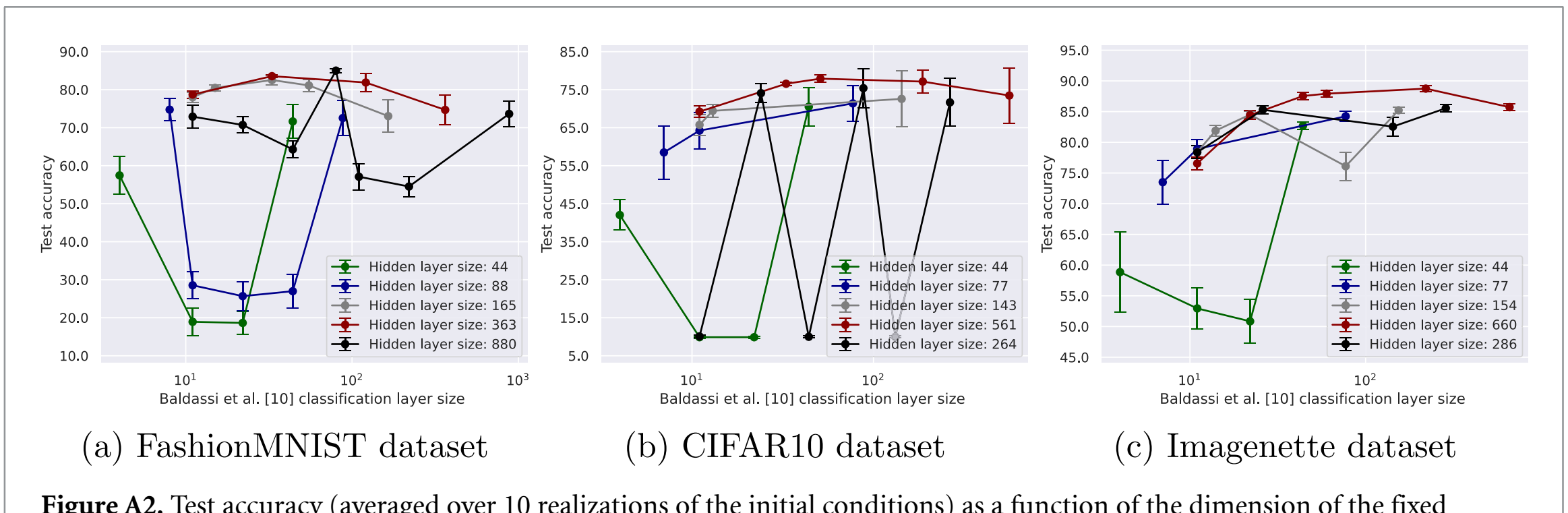

(a) FashionMNIST dataset (b) CIFAR10 dataset (c) Imagenette dataset

**Figure A2.** Test accuracy (averaged over 10 realizations of the initial conditions) as a function of the dimension of the fixed (non-learnable) classification layer used in [10].

### A.3. Adam optimizer

In this section, we compare the results of our proposed solution (reported in section 4.2) with those obtained using full-precision Adam, as shown in figure A3. Specifically, we trained two hidden layer floating-point MLPs using full-precision Adam with the following set of hyperparameters $\mathcal{H}$: batch size $bs = 100$, number of epochs $e = 50$, and learning rate $\eta = 0.001$.

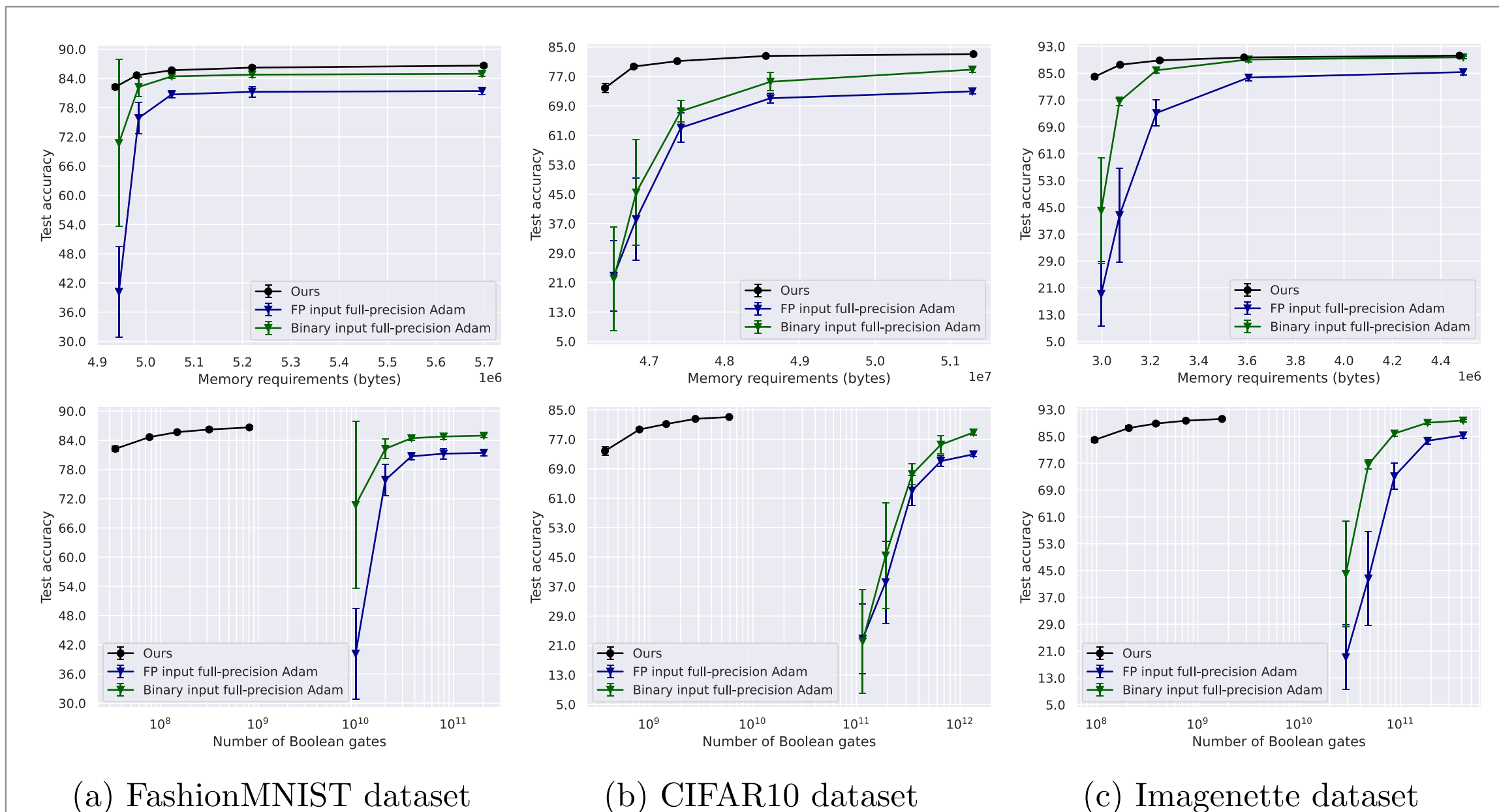

(a) FashionMNIST dataset (b) CIFAR10 dataset (c) Imagenette dataset

**Figure A3.** The results compare our multi-layer fully binary and gradient-free algorithm with full-precision Adam baseline. (Top panels) Test accuracy (averaged over 10 realizations of the initial conditions) as a function of total memory requirements. (Bottom panels) Test accuracy (averaged over 10 realizations of the initial conditions) as a function of the Boolean gate count.

### A.4. Convergence of BNN accuracy

In this section, we evaluate the convergence behavior of our proposed solution. Specifically, figure A4 shows the accuracy progression over epochs for both our method and full-precision SGD, using the larger architectures from the experiments in section 4.2, which have the highest memory requirements.

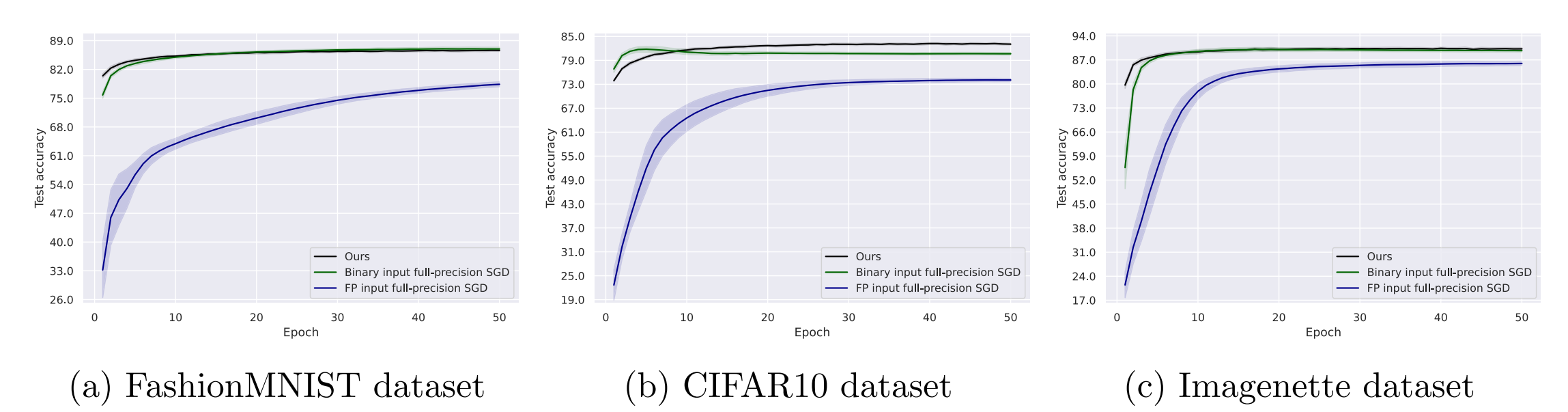

(a) FashionMNIST dataset (b) CIFAR10 dataset (c) Imagenette dataset

**Figure A4.** Test accuracy progression (averaged over 10 realizations of the initial conditions) over epochs. The results compare our multi-layer fully binary and gradient-free algorithm with full-precision SGD baseline, using the larger architectures from the experiments in section 4.2.

### A.5. Evaluating catastrophic forgetting

In this section, we evaluate the catastrophic forgetting issue by constraining the number of bits that the integer-valued hidden weights can take, as shown in figure A5. We conduct this analysis using the medium-sized architectures employed in the experiments of section 4.2, specifically with two hidden layers of size 135. Notably, when the hidden weights are limited to 1 bit, only the visible binary weights are retained, effectively disabling the metaplastic memory mechanism.

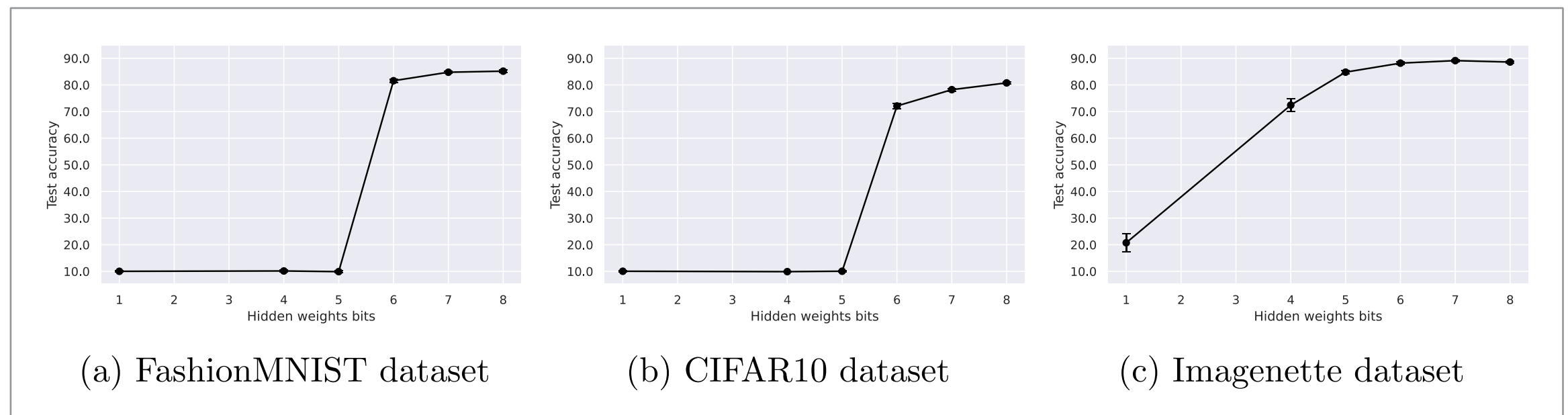

(a) FashionMNIST dataset (b) CIFAR10 dataset (c) Imagenette dataset

**Figure A5.** Test accuracy (averaged over 10 realizations of the initial conditions) as a function of the number of bits available for the integer-valued hidden weights. Results are reported for the medium-sized architectures used in section 4.2, consisting of two hidden layers with 135 neurons each.

### A.6. Training the fixed random local output classifiers

In this section, we evaluate the impact of keeping the local output classifiers randomly initialized and fixed during training. Specifically, table A1 reports the final test accuracy when the local classifiers are kept fixed, compared to when they are jointly trained with the rest of the model.

**Table A1.** Test accuracy (averaged over 10 realizations of the initial conditions) for models with either trained or fixed local output classifiers.

| Dataset | Hidden layer size | Test accuracy | |
|---|---|---|---|
| | | Trained classifiers | Fixed classifiers |
| MNIST | 35,35 | $81.86 \pm 0.36$ | $82.25 \pm 0.51$ |
| | 135,135 | $85.17 \pm 0.42$ | $85.68 \pm 0.34$ |
| | 525,525 | $86.66 \pm 0.30$ | $86.64 \pm 0.29$ |
| CIFAR10 | 35,35 | $73.44 \pm 1.04$ | $73.85 \pm 1.21$ |
| | 135,135 | $81.32 \pm 0.41$ | $81.16 \pm 0.30$ |
| | 525,525 | $83.13 \pm 0.22$ | $83.07 \pm 0.25$ |
| Imagenette | 35,35 | $84.51 \pm 0.70$ | $83.95 \pm 0.75$ |
| | 135,135 | $88.70 \pm 0.57$ | $88.82 \pm 0.45$ |
| | 525,525 | $90.13 \pm 0.24$ | $90.22 \pm 0.38$ |

## Appendix B. Tabular datasets results

### B.1. UCI

The UCI (University of California, Irvine) ML repository [85] is a widely recognized collection of tabular datasets designed for various applications, including classification, regression, and clustering tasks. In this section, we present results comparing our proposed algorithm with the fully binary single-layer SotA algorithm [10] using three different network sizes: Large (two hidden layers of size 135), Medium (two hidden layers of size 55), and Small (two hidden layers of size 15). The classification layer dimensions of the SotA solution [10] are selected to ensure the same total memory demand. The evaluation is conducted on 30 UCI classification datasets, spanning a diverse range of sample sizes, features, and classes, as shown in figure B1. Note that, for all UCI datasets and both algorithms, we used a binarization method based on a 32-bit distributive thermometer encoder [64].

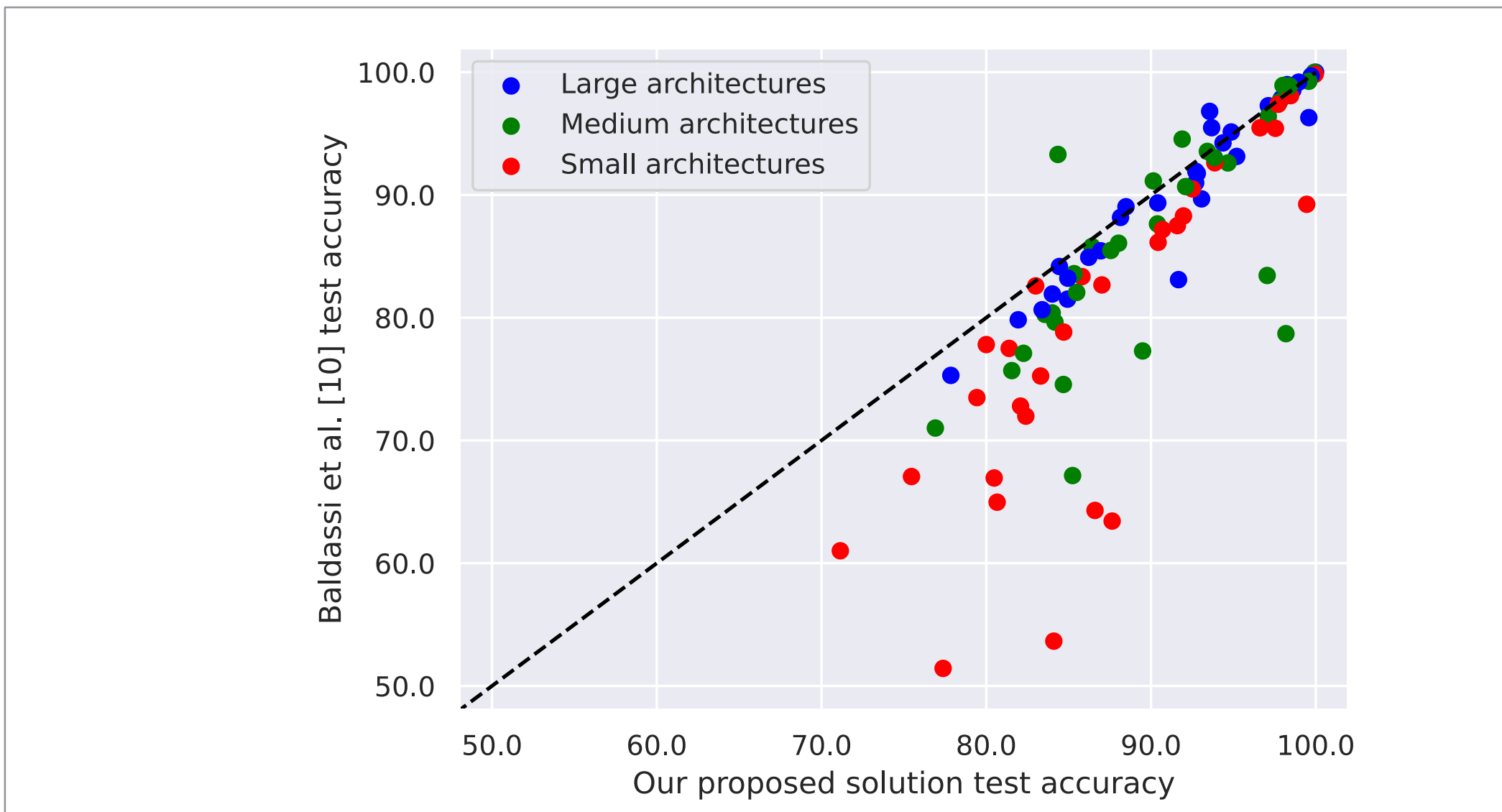

**Figure B1.** Test accuracy of our proposed solution versus test accuracy of the SotA solution [10]. Each point represents a different UCI dataset. Large architectures correspond to BMLPs with two hidden layers of size 135, medium architectures to two hidden layers of size 55, and small architectures to two hidden layers of size 15. The classification layer dimensions of the SotA solution [10] are selected to ensure the same total memory demand. As shown, the smaller the architecture (and consequently the memory demand), the more our algorithm outperforms [10] in test accuracy.

# Appendix C. Tabular format results

## C.1. Comparison with the SotA algorithm [10] and full-precision SGD

In this section, we present the results of section 4.2 in tabular format. Specifically, tables C1–C4 show the results for the Random Prototypes, FashionMNIST, CIFAR10, and Imagenette datasets, respectively.

**Table C1.** Results on the Random Prototypes dataset corresponding to figure 3.

| Algorithm | Hidden layer size | Memory (MB) | Boolean gates ($\times 10^9$) | Test accuracy |
|---|---|---|---|---|
| Ours | 35,35 | 1.29 | 0.04 | 79.10 ± 1.27 |
| Baldassi *et al* [10] | 44 | 1.30 | 0.05 | 43.64 ± 4.95 |
| Binary input full-precision SGD | 5,5 | 1.30 | 20.50 | 56.28 ± 5.99 |
| Ours | 75,75 | 1.35 | 0.10 | 87.65 ± 0.56 |
| Baldassi *et al* [10] | 88 | 1.35 | 0.10 | 61.21 ± 2.53 |
| Binary input full-precision SGD | 10,10 | 1.34 | 41.32 | 85.36 ± 2.39 |
| Ours | 135,135 | 1.43 | 0.18 | 92.25 ± 0.45 |
| Baldassi *et al* [10] | 165 | 1.43 | 0.19 | 74.04 ± 1.88 |
| Binary input full-precision SGD | 20,20 | 1.43 | 83.87 | 91.45 ± 0.84 |
| Ours | 255,255 | 1.62 | 0.38 | 93.60 ± 0.67 |
| Baldassi *et al* [10] | 330 | 1.62 | 0.39 | 84.49 ± 0.91 |
| Binary input full-precision SGD | 41,41 | 1.62 | 177.14 | 93.84 ± 0.58 |
| Ours | 525,525 | 2.16 | 0.94 | 94.83 ± 0.56 |
| Baldassi *et al* [10] | 814 | 2.16 | 1.01 | 90.62 ± 0.73 |
| Binary input full-precision SGD | 98,98 | 2.16 | 457.07 | 94.51 ± 0.60 |

**Table C2.** Results on the FashionMNIST dataset corresponding to figure 4(a).

| Algorithm | Hidden layer size | Memory (MB) | Boolean gates ($\times 10^9$) | Test accuracy |
|---|---|---|---|---|
| Ours | 35,35 | 4.94 | 0.03 | 82.25 ± 0.51 |
| Baldassi *et al* [10] | 44 | 4.94 | 0.04 | 57.45 ± 4.93 |
| FP input full-precision SGD | 5,5 | 4.94 | 16.16 | 53.07 ± 13.50 |
| Binary input full-precision SGD | 5,5 | 4.94 | 16.16 | 81.60 ± 1.26 |
| Ours | 75,75 | 4.98 | 0.08 | 84.66 ± 0.32 |
| Baldassi *et al* [10] | 88 | 4.98 | 0.08 | 74.75 ± 2.96 |
| FP input full-precision SGD | 11,11 | 4.98 | 35.97 | 72.53 ± 4.28 |
| Binary input full-precision SGD | 11,11 | 4.98 | 35.97 | 84.82 ± 0.45 |
| Ours | 135,135 | 5.05 | 0.15 | 85.68 ± 0.34 |
| Baldassi *et al* [10] | 165 | 5.05 | 0.15 | 80.48 ± 0.84 |
| FP input full-precision SGD | 21,21 | 5.06 | 69.96 | 76.84 ± 1.52 |
| Binary input full-precision SGD | 21,21 | 5.06 | 69.96 | 86.21 ± 0.48 |
| Ours | 255,255 | 5.22 | 0.31 | 86.22 ± 0.34 |
| Baldassi *et al* [10] | 363 | 5.23 | 0.34 | 83.56 ± 0.42 |
| FP input full-precision SGD | 44,44 | 5.22 | 152.69 | 78.31 ± 1.49 |
| Binary input full-precision SGD | 44,44 | 5.22 | 152.69 | 86.64 ± 0.37 |
| Ours | 525,525 | 5.70 | 0.81 | 86.64 ± 0.29 |
| Baldassi *et al* [10] | 880 | 5.70 | 0.88 | 85.01 ± 0.56 |
| FP input full-precision SGD | 105,105 | 5.69 | 402.97 | 78.38 ± 0.61 |
| Binary input full-precision SGD | 105,105 | 5.69 | 402.97 | 86.98 ± 0.43 |

**Table C3.** Results on the CIFAR10 dataset corresponding to figure 4(b).

| Algorithm | Hidden layer size | Memory (MB) | Boolean gates ($\times 10^9$) | Test accuracy |
|---|---|---|---|---|
| Ours | 35,35 | 46.41 | 0.37 | $73.85 \pm 1.21$ |
| Baldassi *et al* [10] | 44 | 46.49 | 0.47 | $42.03 \pm 3.98$ |
| FP input full-precision SGD | 5,5 | 46.46 | 185.64 | $52.30 \pm 4.91$ |
| Binary input full-precision SGD | 5,5 | 46.46 | 185.64 | $63.60 \pm 9.53$ |
| Ours | 75,75 | 46.79 | 0.81 | $79.68 \pm 0.47$ |
| Baldassi *et al* [10] | 77 | 46.80 | 0.82 | $58.50 \pm 6.99$ |
| FP input full-precision SGD | 10,10 | 46.83 | 371.61 | $66.61 \pm 1.73$ |
| Binary input full-precision SGD | 10,10 | 46.83 | 371.61 | $77.18 \pm 1.01$ |
| Ours | 135,135 | 47.37 | 1.46 | $81.16 \pm 0.30$ |
| Baldassi *et al* [10] | 143 | 47.41 | 1.52 | $69.43 \pm 1.72$ |
| FP input full-precision SGD | 18,18 | 47.43 | 669.78 | $71.24 \pm 1.14$ |
| Binary input full-precision SGD | 18,18 | 47.43 | 669.78 | $80.08 \pm 0.34$ |
| Ours | 255,255 | 48.55 | 2.79 | $82.57 \pm 0.30$ |
| Baldassi *et al* [10] | 264 | 48.54 | 2.81 | $74.09 \pm 2.46$ |
| FP input full-precision SGD | 33,33 | 48.55 | 1230.92 | $72.68 \pm 0.53$ |
| Binary input full-precision SGD | 33,33 | 48.55 | 1230.92 | $80.57 \pm 0.30$ |
| Ours | 525,525 | 51.31 | 5.91 | $83.07 \pm 0.25$ |
| Baldassi *et al* [10] | 561 | 51.32 | 5.99 | $77.90 \pm 1.00$ |
| FP input full-precision SGD | 70,70 | 51.32 | 2626.65 | $74.08 \pm 0.41$ |
| Binary input full-precision SGD | 70,70 | 51.32 | 2626.65 | $80.66 \pm 0.33$ |

**Table C4.** Results on the Imagenette dataset corresponding to figure 4(c).

| Algorithm | Hidden layer size | Memory MB) | Boolean gates ($\times 10^9$) | Test accuracy |
|---|---|---|---|---|
| Ours | 35,35 | 2.97 | 0.10 | $83.95 \pm 0.75$ |
| Baldassi *et al* [10] | 44 | 2.99 | 0.12 | $58.86 \pm 6.50$ |
| FP input full-precision SGD | 5,5 | 2.98 | 46.71 | $48.65 \pm 6.92$ |
| Binary input full-precision SGD | 5,5 | 2.98 | 46.71 | $74.06 \pm 8.37$ |
| Ours | 75,75 | 3.08 | 0.21 | $87.50 \pm 0.55$ |
| Baldassi *et al* [10] | 77 | 3.07 | 0.21 | $73.51 \pm 3.59$ |
| FP input full-precision SGD | 10,10 | 3.08 | 93.74 | $77.03 \pm 3.10$ |
| Binary input full-precision SGD | 10,10 | 3.08 | 93.74 | $86.79 \pm 0.75$ |
| Ours | 135,135 | 3.24 | 0.38 | $88.82 \pm 0.45$ |
| Baldassi *et al* [10] | 154 | 3.25 | 0.41 | $81.90 \pm 0.89$ |
| FP input full-precision SGD | 19,19 | 3.25 | 179.16 | $83.56 \pm 1.09$ |
| Binary input full-precision SGD | 19,19 | 3.25 | 179.16 | $89.00 \pm 0.67$ |
| Ours | 255,255 | 3.59 | 0.76 | $89.69 \pm 0.40$ |
| Baldassi *et al* [10] | 286 | 3.57 | 0.77 | $85.29 \pm 0.64$ |
| FP input full-precision SGD | 36,36 | 3.58 | 343.17 | $85.00 \pm 0.80$ |
| Binary input full-precision SGD | 36,36 | 3.58 | 343.17 | $89.24 \pm 0.41$ |
| Ours | 525,525 | 4.48 | 1.73 | $90.22 \pm 0.38$ |
| Baldassi *et al* [10] | 660 | 4.48 | 1.80 | $87.92 \pm 0.50$ |
| FP input full-precision SGD | 81,81 | 4.47 | 794.10 | $85.96 \pm 0.66$ |
| Binary input full-precision SGD | 81,81 | 4.47 | 794.10 | $89.70 \pm 0.40$ |

## ORCID iDs

Luca Colombo 0000-0002-1986-6459
Fabrizio Pittorino 0000-0002-1919-6141
Manuel Roveri 0000-0001-7828-7687